\documentclass[]{fairmeta}
\usepackage[utf8]{inputenc} 
\usepackage[T1]{fontenc}    
\usepackage{hyperref}       
\usepackage{url}            
\usepackage{booktabs}       
\usepackage{amsfonts}       
\usepackage{nicefrac}       
\usepackage{microtype}      
\usepackage{xcolor}         
\usepackage{longtable}
\usepackage{booktabs}
\usepackage{multirow}
\usepackage{rotating}
\usepackage{makecell}
\usepackage{amsmath}
\newcommand{\easy}{\textcolor{blue}}
\newcommand{\medium}{\textcolor{orange}}
\newcommand{\hard}{\textcolor{red}}
\usepackage[square,numbers]{natbib}

\title{IntPhys 2: Benchmarking Intuitive Physics Understanding In Complex Synthetic Environments}

\author[1]{Florian Bordes}
\author[1]{Quentin Garrido}
\author[1]{Justine T Kao}
\author[1]{Adina Williams}
\author[1]{Michael Rabbat}
\author[1]{Emmanuel Dupoux}

\affiliation[1]{FAIR at Meta}

\abstract{We present IntPhys 2, a video benchmark designed to evaluate the intuitive physics understanding of deep learning models. Building on the original IntPhys benchmark~\citep{riochet2018intphys}, IntPhys 2 focuses on four core principles related to  macroscopic objects: Permanence, Immutability, Spatio-Temporal Continuity, and Solidity. These conditions are inspired by research into intuitive physical understanding emerging during early childhood. IntPhys 2 offers a comprehensive suite of tests, based on the violation of expectation framework, that challenge models to differentiate between possible and impossible events within controlled and diverse virtual environments. Alongside the benchmark, we provide performance evaluations of several state-of-the-art models. Our findings indicate that while these models demonstrate basic visual understanding, they face significant challenges in grasping intuitive physics across the four principles in complex scenes, with most models performing at chance levels (50\%), in stark contrast to human performance, which achieves near-perfect accuracy. This underscores the gap between current models and human-like intuitive physics understanding, highlighting the need for advancements in model architectures and training methodologies.
}

\date{\today}
\correspondence{Florian Bordes at \email{fbordes@meta.com}}

\metadata[Code]{\url{https://github.com/facebookresearch/IntPhys2}}
\metadata[Blogpost]{\url{https://ai.meta.com/blog/v-jepa-2-world-model-benchmarks}}

\begin{document}

\maketitle

\section{Introduction}
\label{section:intro}
Understanding intuitive physics is a fundamental aspect of human cognition~\citep{piaget1954construction,baillargeon_permanence_1991,baillargeon_support_1992,baillargeon_support_1990,spelke_spatiotemporal_1995}, enabling individuals to effectively navigate and interact with the physical world. In recent years, there has been a growing interest in replicating this intuitive understanding within artificial systems~\citep{battaglia_simulation_2013,watters_visual_interaction_networks_2017,piloto_intuitive_2022,riochet2020occlusion}. However, despite advances in machine learning and computer vision, current models still fall short of human capabilities in this domain~\citep{riochet2018intphys,weihs2022benchmarking,jassim_grasp_2024,bisk2020piqa,benchekroun2023worldsense,bansal2024videophy,bear2021physion}. The IntPhys benchmark \citep{riochet2018intphys} was originally introduced to address the challenge of evaluating intuitive physics understanding in AI models, providing a standardized framework for assessment. However, the benchmark had some limitations, focusing on simple environments that lacked the variations and complexities found in the real world. Furthermore, recent work \citep{garrido2025intuitive} has shown that the benchmark has become saturated, with predictive models such as V-JEPA\citep{bardes_vjepa_2024} achieving high performance on it, highlighting the need for a more challenging and diverse intuitive physics benchmark.

In this paper, we present IntPhys 2, an expanded and more comprehensive benchmark designed to push the boundaries of intuitive physics understanding. IntPhys 2 evaluates four key conditions inspired by human cognition: Object Permanence~\cite{baillargeon_permanence_1991}, Object Immutability~\cite{wilcox1999constancy,wilcox2004constancy}, Spatio-Temporal Continuity~\cite{spelke_origins_1992}, and Solidity~\cite{spelke_origins_1992}. These conditions are carefully selected to encompass a broad range of physical principles, thereby providing a rigorous assessment of model capabilities.
The dataset contains 1416 videos that are divided in 3 different splits. The videos in the Debug and Main splits are released along with their respective metadata while the last split is an Held Out set, in which we release only the videos to avoid training data contamination. Unlike its predecessor that contained very basic and not fully realistic scenes, IntPhys~2 utilizes the full potential of Unreal Engine,\footnote{As a reminder, any use of content or technologies made available by Unreal and/or Epic Games, or any other provider, should comply with their applicable terms (such as the Content License Agreement available at \url{https://www.unrealengine.com/en-US/eula/content} or any other direct agreement one may have with Epic / Unreal)} using photorealistic environments made with dynamic shadows and lighting to better simulate real-world settings. IntPhys~2 improves upon the original IntPhys benchmark by introducing more realistic occlusions through the use of both fixed and moving cameras. Movement-based occlusions are more natural, capturing situations such as those that occur when an observer moves their head to look away and then back to the original point of view. By incorporating both fixed and moving cameras as well as using more complex scenes, IntPhys~2 provides a more comprehensive evaluation framework for intuitive physics understanding.

Using IntPhys 2, we performed a comprehensive performance evaluations of state-of-the-art predictive models and Multimodal Large Language Models (MLLMs)\citep{bordes2024introductionvisionlanguagemodeling}. While these models have achieved notable advancements, our findings indicate that they continue to struggle with the nuances of simple intuitive physics properties such as permanence and immutability, particularly in comparison to human performance, which remains consistently strong across all conditions. This disparity highlights the ongoing challenges in bridging the gap between artificial and human cognition, emphasizing the need for continued research and innovation in this critical area. 

Our key contributions are as follows:
\begin{itemize}
\item A novel video benchmark dataset for intuitive physics, featuring diverse scenarios with varying complexity levels. IntPhys 2 advances beyond existing benchmarks by incorporating photorealistic scenes with sophisticated visual elements (including complex lighting, shadows, occlusions, and textures), and employing both fixed and dynamic camera perspectives to simulate natural viewpoint changes.
\item A comprehensive evaluation of state-of-the-art AI systems, including predictive models and Multimodal Large Language Models (MLLMs), establishing new baselines and identifying specific challenges in intuitive physics reasoning.
\end{itemize}

\begin{figure}
    \centering
    \includegraphics[width=1.0\linewidth]{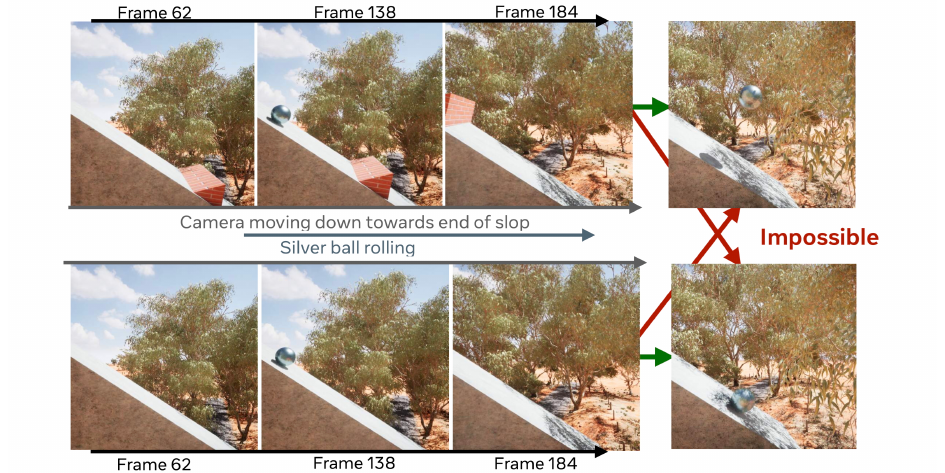}
    \caption{Example of a scene in IntPhys2, which follows a similar design to IntPhys1. Each scene consists of a set of four videos. Two pairs depict possible outcomes, while the other two represent impossible outcomes. The presence of an obstacle or occluder determines the outcome: a possible outcome in the first pair becomes impossible in the second, and vice versa. In this example, a silver ball rolls down a path. If a brick obstacle is present, the ball should collide with it and change its trajectory. If the ball passes through the brick obstacle without altering its path, this outcome is deemed impossible. Conversely, when no obstacle is present, the ball's trajectory should remain unchanged, making this the likely outcome. }
    \label{fig:Prompt_sensitivity}
\end{figure}

\section{Benchmark Design}
\label{section:design}

From an early age, humans develop an innate ability to grasp basic physical principles~\citep{piaget1954construction,baillargeon_permanence_1991,baillargeon_support_1992,baillargeon_support_1990,spelke_spatiotemporal_1995}, such as object permanence~\citep{baillargeon_permanence_1991} (objects persist in space and time, even when they are out of sight), immutability~\cite{wilcox1999constancy,wilcox2004constancy} (objects maintain their shape and structure), spatio-temporal continuity~\cite{spelke_origins_1992} (objects move smoothly through space and time), and solidity~\cite{spelke_origins_1992} (objects occupy space and cannot pass through one another). These principles allow us to predict and interpret the behavior of inanimate objects in our environment, forming the foundation for more complex reasoning and decision-making processes. To systematically assess the development of these intuitive physics principles, the violation of expectations (VOE)~\cite{margoni_voe_2024,spelke1985voe_looking} paradigm has been leveraged. This paradigm, which has been extensively used in studies with human infants\citep{baillargeon1985object, Spelke1995}, involves presenting them with scenarios where objects either behave in accordance with or violate these fundamental physical principles. By measuring infants' gaze time to these scenarios, researchers can infer their understanding of intuitive physics. Such a framework has been one of the main inspirations for IntPhys~\citep{riochet2018intphys}, and IntPhys 2 builds upon this foundation by adhering to the methodological framework established by its predecessor, employing a quadruplet video structure for each scene. This design comprises two possible and two impossible videos per scenario, configured such that the possible video of one scenario serves as the impossible video in another, and vice versa. This systematic arrangement is instrumental in mitigating low-level perceptual biases, thereby requiring models to engage with high-level temporal dependencies and underlying physical principles. By maintaining this rigorous structure, IntPhys 2 offers a robust and unbiased framework for assessing the depth of intuitive physics understanding in machine learning systems, preventing models from relying on shortcuts or latching onto spurious features and ensuring that model performance is more correlated with genuine cognitive capabilities rather than the exploitation of dataset-specific artifacts.

\begin{figure}
    \centering
    \includegraphics[width=1.0\linewidth]{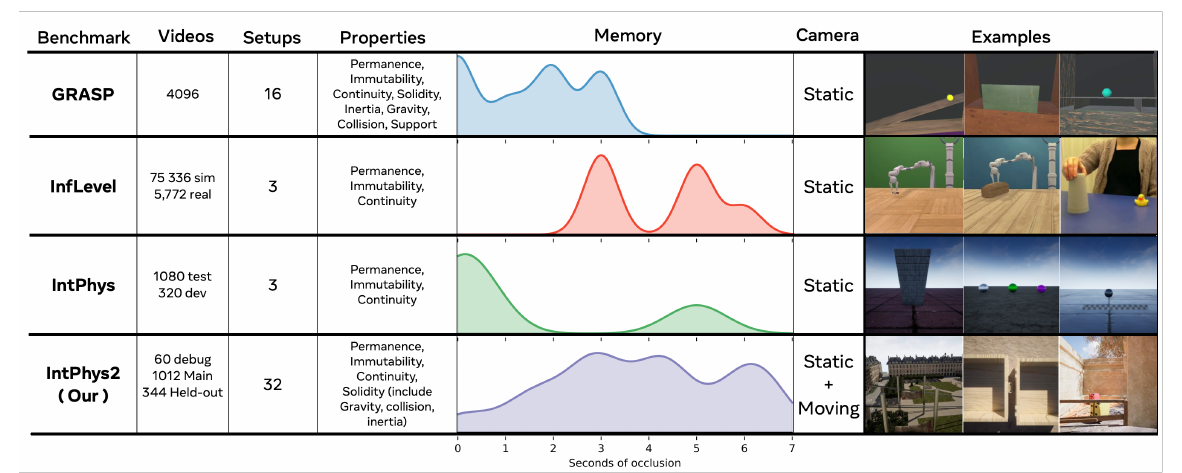}
    \caption{\textbf{Benchmark Comparison.} Our analysis compares four benchmarks: GRASP~\cite{jassim_grasp_2024}, InfLevel~\cite{weihs2022benchmarking}, IntPhys~\cite{riochet2018intphys}, and IntPhys2 (Ours). The benchmarks differ in their number of videos (simulated, real, test, development, debug, main, held-out), experimental setups, and physical properties assessed (permanence, immutability, continuity, solidity, inertia, gravity, collision, support). The density plots illustrate the distribution of occlusion durations (in seconds) for each benchmark. In contrast to other benchmarks, IntPhys2 covers a higher range of occlusion durations, allowing for a better assessment of a model's short-term memory. Camera settings vary between static and moving configurations. Example frames from each benchmark are shown on the right.}
    \label{fig:comp-bench}
\end{figure}

IntPhys 2 introduces several key advancements over previous benchmarks and datasets in the domain of intuitive physics understanding~\cite{jassim_grasp_2024,weihs2022benchmarking,riochet2018intphys} that are illustrated in Figure \ref{fig:comp-bench}. These enhancements are designed to provide a more rigorous and comprehensive evaluation of AI models, addressing limitations observed in earlier works. The core differences are as follows:
\begin{itemize}
    \item \textbf{Focus on Occlusions}: Unlike previous benchmarks that may have included a variety of scenarios, IntPhys 2 exclusively considers occlusions. This focus allows for a more targeted assessment of a model's ability to maintain their understanding in the presence of visual obstructions.
    \item \textbf{Dynamic Camera Movements}: To create occlusions, IntPhys 2 employs static and dynamic camera movements. This approach not only increases the complexity of the scenes but also mimics real-world conditions where objects may be temporarily obscured from view due to changes in perspective.
    \item \textbf{Enhanced Realism}: The scenes in IntPhys 2 are crafted with improved realism, providing a more lifelike and challenging environment for models to navigate. This enhancement ensures that the benchmark more accurately reflects the complexities of the real-world.
    \item \textbf{Diverse Scene Variety}: IntPhys 2 significantly expands the diversity of scenes and tasks considered. Unlike traditional datasets that often feature a single scene per physical property, IntPhys 2 includes multiple tasks within each condition, offering a broader range of challenges and reducing the risk of overfitting to specific scene types.
    \item \textbf{Increased Short-Term Memory Demand}: The benchmark places a stronger emphasis on the need for short-term memory, requiring models to retain and utilize information over brief intervals effectively. This demand is critical for accurately predicting and understanding the dynamics of occluded objects.
\end{itemize}

\begin{table}[ht]
\centering
\caption{{\bf IntPhys2 benchmark splits.} We release three separate splits. The first is intended for debugging only and provide some measurement on the model's sensitivity to the video generation artifacts (such as mp4 compression or cloud moving the background of the scene). The second is the main evaluation set with three different sub-splits ("Easy", "Medium", "Hard"). The third is a held-out split that we release without additional metadata.}
\label{tab:benchmark_splits}
\scriptsize
\begin{tabular}{lllll}
\toprule
\textbf{Split} & \textbf{Scenes} & \textbf{Videos} & \textbf{Description} & \textbf{Purpose} \\
\midrule
\textbf{Debug Set} & 5 & 60 & Static cameras, bright assets, 3 generations & Model calibration \\
\midrule
\textbf{Main Set} & 253 & 1,012 & Static and moving cameras: 3 sub-splits: & Main evaluation set \\
& & & - Easy: Simple environments, colorful shapes & \\
& & & - Medium: Diverse backgrounds, textured shapes & \\
& & & - Hard: Realistic objects, complex backgrounds & \\
\midrule
\textbf{Held-Out Set} & 86 & 344 & Moving cameras, Mirrors hard sub-split, includes distractors & Main test set \\
\bottomrule
\end{tabular}
\label{tab:intphys2_splits}
\end{table}

We designed the benchmark with three distinct data splits to facilitate comprehensive evaluation (Table \ref{tab:intphys2_splits}). The first split, known as the Debug set, includes five scenes with static camera setups and brightly colored assets. Each scene features a quadruplet of videos, supplemented by three additional videos that, while identical in configuration, display subtle variations due to environmental factors such as cloud movement or wind. These variations, though easy to miss to human observers, introduce minor pixel-level discrepancies that can influence model performance. This split is primarily intended for model calibration and to evaluate sensitivity to such noise. Ideally, a model should be robust to these negligible variations, demonstrating its ability to generalize beyond minor environmental fluctuations. We demonstrate ways to use this subset for qualitative analysis of predictive models in appendix~\ref{sec:qual-debug}. The second split, termed the main set, comprises 253 scenes resulting in a total of 1,012 videos. This set is designed for zero-shot evaluations and serves as the primary basis for comparing models. It is further divided into three sub-splits: an easy sub-split with simple environments and shape-based colorful assets, a medium sub-split featuring more diverse backgrounds and simple shapes with complex textures, and a hard sub-split characterized by highly diverse backgrounds and assets shaped like real objects. The final split, is a held-out set for which metadata (containing the ground truth about the plausibility of the events in the video) is not released. Solving this hard subset requires an understanding that is robust to complex settings and that does not use metadata as additional help. While the main set allows one to track progress more granularly and can be more lenient on models, solving the held-out set is what will truly mean that a model understands intuitive physics in complex settings.

\section{Evaluation protocol : measuring intuitive physics understanding}
\label{section:baseline_eval}

\paragraph{\textbf{Human Evaluation.}} To assess the gap between human and AI performance in understanding intuitive physics, we conducted a human annotation task. This task was designed to evaluate the ability of human participants to judge the physical plausibility of videos generated by a simulation engine, similar to those used in video game development. The evaluation involved the 1,416 videos, each with a duration of around 10 seconds. To ensure a comprehensive assessment, each video was rated by three different annotators. The order of video presentation was randomized for each annotator to mitigate order effects and maintain the integrity of the evaluation process. To prevent attention fatigue and maintain high-quality ratings, each annotator was limited to evaluating a maximum of 96 videos. Prior to the main annotation task, annotators were shown only a set of 10 videos that were all physically plausible. This familiarization phase was designed to acquaint annotators with the types of videos produced by the simulation engine and to establish a baseline understanding of physical plausibility within the context of the task. Annotators were only informed that the initial set of 10 videos they saw depicted scenes where objects behaved in physically plausible ways. Following this, annotators were tasked with evaluating additional videos, some of which might contained errors that resulted in physically implausible object behavior. Annotators were instructed to watch each video carefully and in its entirety, considering the plausibility of object behavior based on real-world physics. They rated each video using a Likert scale, ranging from 1 (completely implausible) to 4 (very plausible). Then, we aggregated the results using a majority vote between the three annotators. This structured approach to human evaluation was crucial for obtaining reliable data on human performance, which serves as a benchmark for comparing the capabilities of current and future AI models in understanding intuitive physics.

\paragraph{\textbf{Evaluating multimodal large language models.}}
Our evaluation methodology for MLLMs diverges slightly from our human evaluation since current models 1) are not yet able to process 10 videos in their input context 2) do not have long term visual memory 3) are not learning from previous context. To compensate for this lack of memory, we employed more detailed prompts for MLLMs asking only wether the video despite a plausible scenario. The prompts included explicit instructions about the video source being a simulator and that the model should base its answer solely on the events happening in the video, not on the quality of the simulation itself. To assess the models' sensitivity to prompts, we evaluated each model point-wise using the prompts presented in Table \ref{tab:prompts}. The first prompt is concise and open-ended, requiring the model to respond with a simple "yes" or "no". In contrast, the second prompt is more specific, guided, and is expecting a binary digit as response. Anecdotally, MLLMs can be sensitive to the format of the requires output\citep{long2025llmsbiasedoutputformats}, so we made a version of the second prompt in which we require a "yes/no" answer instead of the binary digit format. However, prompting is not the sole source of variance; sometimes, even with a temperature setting of 0, models can produce different answers to the same prompt and input data. Therefore, we ran each prompt at least twice to evaluate any variance in predictions. Ideally, the accuracy should remain consistent in such cases. To give MLLMs an advantage to compensate for their short-term memory limitations, we decided to show the best accuracy that can be obtained by a model across multiple different runs, instead of doing a majority vote like we did in the human evaluation. Lastly, since the number of frames that can be fed into an MLLM depends on its input context size, we had to run several experiments using a different number of input frames.

\paragraph{\textbf{Evaluating prediction-based models.}}
Inspired by the Violation of Expectation framework, \citet{garrido2025intuitive} introduced a model-based evaluation setup that measures how much a model is surprised when viewing an unexpected event compared to an expected event. A higher surprise indicates that the events violated the model's expectations, with impossible videos expected to elicit more surprise than possible ones. A proxy for surprise is the prediction error over a video for models that can predict the future~\citep{riochet2018intphys,smith_adept_2019,riochet2020occlusion,garrido2025intuitive}. We split a video into overlapping windows (typically 16-32 frames) that the model can process. For each window, the model predicts the target part based on the context, and the prediction error measures the model's surprise. Comparing surprise across videos probes the model's understanding, and we adapt the protocol for longer contexts as described in appendix~\ref{sec:pred-protocol}. The adaptation introduces constraints: events necessary for prediction must be in the context, and models must remember occluded objects. Models handling 16 frames at a time require a suitable framerate for prediction in order to balance memory and motion fluidity. Different experimental settings have different baseline prediction errors, making surprise comparisons challenging. Paired videos with identical content except for a physics-breaking event allow for controlled comparisons as the surprise difference can be attributed to the physics-breaking event, enabling precise probing of physics understanding, and giving rise to two evaluation protocols: pairwise and single video settings.

While we have described protocols designed for certain families of models, playing to their respective strengths, other protocols are possible. For example, in InfLevel-lab~\cite{weihs2022benchmarking}, models trained for action recognition are probed by measuring how out of distribution impossible videos are. This has however not yielded evidence of understanding in models. We thus chose to focus on models that have either demonstrated previous understanding~\cite{garrido2025intuitive} or ones that can be probed akin to humans.

\section{Experiments}

In our evaluation, we investigated the capabilities of several state-of-the-art MLLMs, including both open-source and proprietary options. Our study featured the Gemini series (Gemini 1.5 Pro and Gemini 2.5 Pro Flash Preview \citep{geminiteam2024geminifamilyhighlycapable}), as well as the latest versions of GPT4-o and Qwen-VL 2.5 \citep{Qwen2.5-VL}. We also evaluated four prediction-based methods: VideoMAEv2~\cite{wang_videomaev2_2023} which predicts pixels directly, Cosmos-Predict1-4B~\cite{agarwal2025cosmos} which predicts in the latent space of an autoencoder, as well as V-JEPA~\cite{bardes_vjepa_2024} and V-JEPA 2~\cite{vjepa_2} which predicts in latent space.
The main results are presented in Table \ref{tab:accuracy}. Notably, there is a meaningful gap between human and current model performances. The best MLLM, Gemini 2.5 Flash, performed only slightly above random chance, except on the easy subset of our benchmark, where it achieved 64\% accuracy. Similarly for predictive models, V-JEPA 2 achieves the highest performance, yet remains below 60\%. Table \ref{tab:accuracy_detail} provides more fine-grained results across four different conditions. The permanence condition appears to be the easiest for both models and humans, as objects are not moving by themselves. While there is no consistent trend for fixed versus moving camera scenarios among models, humans tend to perform slightly better in fixed camera settings. Overall, the gap between human and model performance remains significant across each split and data category.

\begin{table}[ht]
\centering
\footnotesize
\caption{\textbf{Accuracy values showing best model performance for each difficulty level.} Most of the models were run a dozen of time with a different set of hyper-parameters. For a given model, we only report its best run for a given column (except the Held Out which was run only one time with the best hyper-parameters found for overall).The human performance is computed from a majority vote between the annotators.}
\begin{tabular}{l|c|ccc|c||c|c}
\toprule
Model & Type & Easy & Medium & Hard & Overall & Held Out & \textit{\textcolor{gray}{IntPhys\cite{riochet2018intphys}}}\\
\midrule
GPT4-o\citep{openai2023gpt4} & MLLMs & 57.69 & 54.75 & 54.17 & 53.75 & 53.19 & \textit{\textcolor{gray}{-}} \\
Qwen-VL 2.5\citep{Qwen2.5-VL} & MLLMs & 50.96 & 53.25 & 51.49 & 52.27 & 49.12 & \textit{\textcolor{gray}{-}} \\
Gemini-1.5 Pro\citep{geminiteam2024geminifamilyhighlycapable} & MLLMs & 58.65 & 53.0 & 52.67 & 52.27 & 52.10 & \textit{\textcolor{gray}{55.81}}\\
Gemini-2.5 Flash\citep{geminiteam2024geminifamilyhighlycapable} & MLLMs & \textbf{64.42} & \textbf{56.75} & \textbf{54.46} & \textbf{55.63} & \textbf{56.10} & \textit{\textcolor{gray}{56.39}}\\
\midrule
VideoMAEv2-g~\cite{wang_videomaev2_2023} & Predictive & 46.00 & \textbf{58.50} & 52.73 & 53.75 & 53.49 & \textit{\textcolor{gray}{59.40}}  \\
Cosmos-4B~\cite{agarwal2025cosmos} & Predictive & 46.00 & 52.00 & 48.05 & 49.41 & 48.84 & \textit{\textcolor{gray}{85.42}}\\
V-JEPA-h + RoPE~\cite{bardes_vjepa_2024,garrido2025intuitive} & Predictive & 52.00 & 53.00 & 57.42 & 53.75 & 54.65  & \textit{\textcolor{gray}{98.30}}\\
V-JEPA 2-h ~\cite{vjepa_2} & Predictive & \textbf{54.00} & \textbf{58.50} & \textbf{59.38} &  \textbf{57.51} & \textbf{56.40} & \textit{\textcolor{gray}{87.22}}\\
\midrule
Human & - & 96.17 & 97.8 & 95.5 & 96.44 & 92.44 & \textit{\textcolor{gray}{-}} \\
\bottomrule
\end{tabular}

\label{tab:accuracy}
\end{table}

\begin{table}[ht]
\centering
\footnotesize
\caption{\textbf{Accuracy across property and camera type.} For each model report the accuracy of each subset based on the best one across a set of hyperparameters. The smaller size of each subsets contributes to volatility in performance.}
\label{tab:category_results}
\begin{tabular}{l|{c}{c}|{c}{c}|{c}{c}|{c}{c}}
\toprule
 & \multicolumn{2}{c|}{Permanence} & \multicolumn{2}{c|}{Immutability} & \multicolumn{2}{c|}{Continuity} & \multicolumn{2}{c}{Solidity} \\
Model & Fixed & Moving & Fixed & Moving & Fixed & Moving & Fixed & Moving \\
\midrule
GPT4-o & 59.62 & \textbf{58.82} & 58.65 & 59.56 & \textbf{54.81} & \textbf{57.35} & \textbf{56.73} & 55.32 \\
Qwen-VL 2.5 & 53.85 & 54.41 & 56.73 & 53.68 & 52.88 & 54.41 & 50.96 & 51.06 \\
Gemini-1.5 Pro & 55.77 & 55.88 & 56.73 & 56.73 & 54.80 & 54.80 & 56.73 & 56.73 \\
Gemini-2.5 Flash & \textbf{64.42} & \textbf{58.82} & \textbf{59.62} & \textbf{63.97} & \textbf{54.81} & 55.15 & 55.77 & \textbf{56.38} \\
\midrule
VideoMAEv2-g~\cite{wang_videomaev2_2023} & \textbf{63.46} & 50.00 & 54.81 & 53.69 & \textbf{65.38} & 54.41 & 48.08 & \textbf{59.57}\\
Cosmos-4B~\cite{agarwal2025cosmos} & 51.92 & 41.18 & 50.96 & 48.32 & 53.85 & 50.00 & 48.08 & 55.32\\
V-JEPA-h + RoPE~\cite{bardes_vjepa_2024,garrido2025intuitive} & \textbf{63.46} & \textbf{67.65} & 51.92 & 56.38 & 50.00 & 57.35 & \textbf{50.00} & 52.13  \\
V-JEPA 2-h~\cite{vjepa_2} & 59.62 & 57.35 & \textbf{55.77} & \textbf{58.72} & 57.69 & \textbf{75.00} & 46.15 & 58.51 \\
\midrule
Human & 100.0 & 99.26 & 97.11 & 90.44 & 99.04 & 94.44 & 96.15 & 95.21 \\
\bottomrule
\end{tabular}
\label{tab:accuracy_detail}
\end{table}

\subsection{Results: Multimodal Large Language Models}

\begin{figure}
\centering
\includegraphics[width=1.0\linewidth]{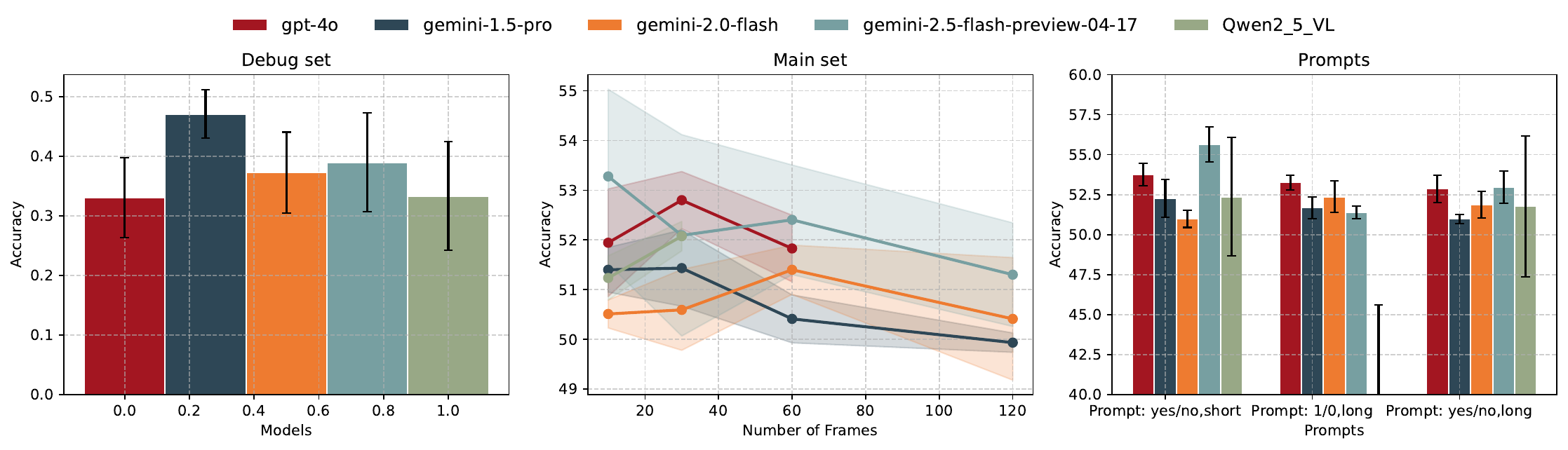}
\caption{\textbf{Evaluation of model's sensitivity.} (left) We conducted an ablation study examining various factors, including sensitivity to different prompts and the model's variability in responses to identical inputs, as well as the difficulty level of the data. (right) We illustrate how a model's performance varies with the number of frames it receives. Our findings indicate that most models struggle to effectively make use of an increased number of input frames.}
\label{fig:NumberOfFrames}
\end{figure}

A key differentiator among these models is their language component. To assess the sensitivity of these models to the prompt, we performed an ablation study over the different prompts described in Table \ref{tab:prompts}. Another key element is the method of processing video input and how many frames the model is fed with. The Gemini models are designed to accept MP4 videos directly, whereas the other models require video content to be converted into sequences of image frames. This required us to adopt a customized approach for each model: for those unable to process MP4s directly, we employed uniform subsampling of frames according to the model's input capacity. Conversely, for the Gemini models, we extended the video length to ensure the model received an adequate number of frames, as their API subsamples video at a rate of 1 frame per second. This allows us to run our ablations using either 10 frames, 30 frames, 60 frames, or 120 frames. Our evaluation included a series of ablation studies, which assessed the impact of various prompts, the number of frames inputted into each model, performance across different data difficulty levels, and sensitivity to randomness. To ensure consistency, we maintained a temperature setting of 0 across all models during testing.
The results of our evaluation are presented in Table \ref{tab:accuracy} and \ref{tab:accuracy_detail}, showcasing the optimal performance outcomes for each model across the various factors we examined.
In Figure \ref{fig:NumberOfFrames}, we present an ablation analysis focusing on several key factors: robustness to generation artifacts, number of frames, and prompt. The first plot on the left utilizes the Debug set in IntPhys2, which contains three videos of the exact same scene. Even if these videos appear identical to humans, models can be very sensitive to any compression artifact noise. The model is considered correct only if it gives the correct answer for all 3 videos generated from the same scene. Thus, this accuracy shows the model's performance as well as its robustness to imperceptible noise.
The second plot in Figure \ref{fig:NumberOfFrames} provides insights into how the number of frames influences model performance. Our analysis reveals that most models experience a decline in performance when additional frames are introduced, suggesting a limitation in handling extended contexts\citep{kuratov2024babilongtestinglimitsllms}. The last plot showcases a prompt ablation, in which we can clearly see that the prompt selected can have a huge impact on performance. The best prompt for one model might not be the best one for another. Interestingly, it seems that models like gemini 2.5-flash perform better when being asked to answer by a yes/no answer than a binary digit. Qwen2.5-VL is not even able to follows the instruction correctly for the 0/1 prompt. These evaluations highlight the difficulties in making fair evaluations of MLLMs, as a given choice of hyperparameters can significantly change their performance. However, even the best models are still close to random performance, highlighting that current models have not learned a good physical world model, which might result in higher variance due to the randomness of the answers.

\subsection{Results: Predictive models}

We evaluate four prediction-based methods - two aimed at predicting pixels, two predicting in a latent space. For the former category, we use Cosmos-Predict-4B~\cite{agarwal2025cosmos} and VideoMAEv2~\cite{wang_videomaev2_2023}. While the prediction of Cosmos happens in a latent space, it is an autoencoder aimed at generation, which is why we classify it as pixel prediction. Furthermore, while Cosmos is directly trained to predict the future, acting as a world model, VideoMAEv2 is trained to reconstruct masked tokens from a video. Using it to predict the future is thus different from its training objective, as is the case for the methods we discuss next. The latent prediction methods we evaluate are the V-JEPA + RoPE~\cite{bardes_vjepa_2024,su2021rope} models trained in~\cite{garrido2025intuitive}, as well as V-JEPA 2~\cite{vjepa_2}.
\begin{figure}[!tbp]
    \centering
    \includegraphics[width=0.32\linewidth]{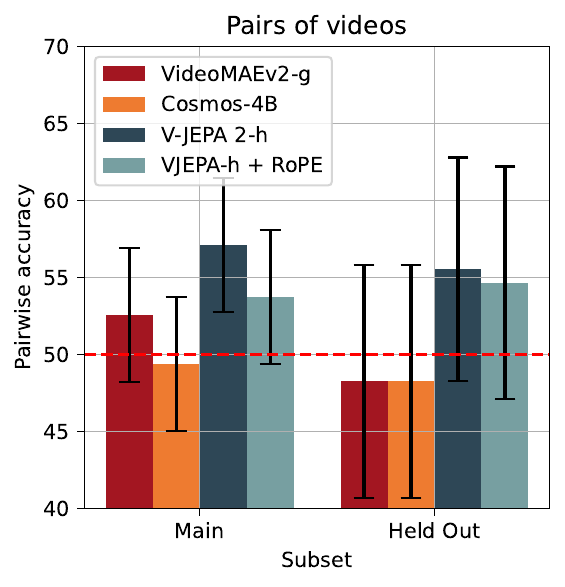}
    \includegraphics[width=0.33\linewidth]{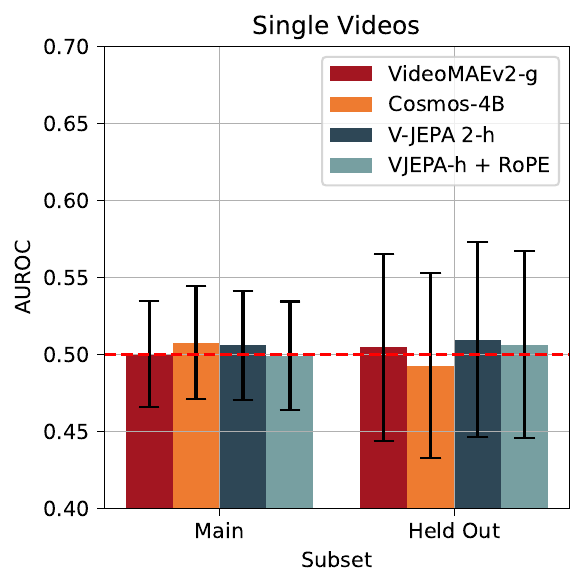}
    \includegraphics[width=0.32\linewidth]{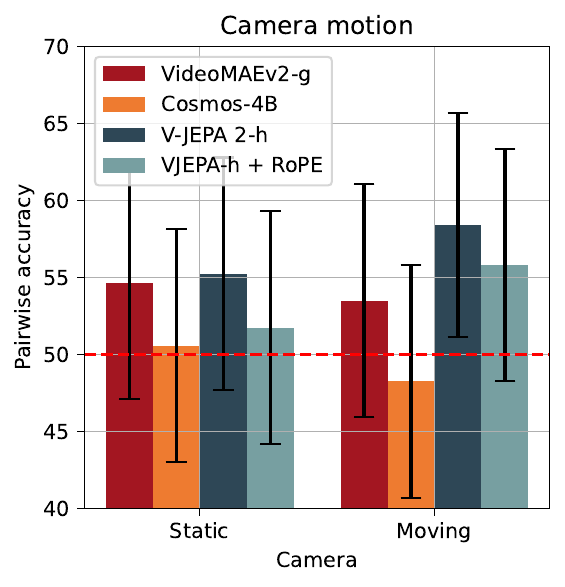}
    \caption{\textbf{Results for predictive models.} (left) When measuring whether models exhibit a higher surprise for impossible instances within a pair, we find that all tested models perform around chance level. (middle) This translates to the harder setting of single video classification, where the performance remains around chance. (right) Focusing on camera movements, one of the key chances in IntPhys 2, we find that model also struggle across camera settings. Confidence intervals obtained via bootstrapping.}
    \label{fig:pred_results}
\end{figure}

We report the accuracy of models on the different subsets of IntPhys 2 in Table~\ref{tab:accuracy}. We find that models exhibit performance close to chance (50\%) across all subsets , as well on the held-out set. This contrasts with the high accuracies that models obtain on IntPhys.
Looking at per condition accuracy in Table~\ref{tab:accuracy_detail}, we further find that even when using more specific subsets of IntPhys 2---and thus using more specialized hyperparameters---the models struggle to surpass chance level. 
We find some exceptions on permanence and continuity, with models achieving above 65\% accuracy. These results should be contextualized against the small size of our subsets, which may introduce randomness in performance, meaning subset results should be taken with a grain of salt.
In Figure~\ref{fig:pred_results}, we further investigate the accuracy of models when classifying pairs or single videos. The latter is conceptually harder as the surprise must be as independent as possible from the general prediction difficulty of the video. It is however how humans are evaluated. We can see in the left and middle of Figure~\ref{fig:pred_results} that while the accuracy can be slightly higher than chance for pairs of video, it remains closer to chance level performance in single videos.\
In  the right of Figure~\ref{fig:pred_results} we ablate further on models' performance on the main subset of IntPhys 2. While the performance on the main set remains low overall, it is possible that the models can perform better on certain subsets. This has been observed on InfLevel-lab, where models are able to perform well on one task, even when they perform close to random chance on others~\cite{garrido2025intuitive}. We thus isolate camera motion and find that even in the fixed camera setting---closest to existing benchmarks---the models perform close to chance level. Interestingly, V-JEPA and V-JEPA 2 perform better in the moving camera setting. This may be explained by slightly different testbeds for both conditions.
These results demonstrate that models still struggle to understand intuitive physical concepts in complex scenarios, even if they are able to demonstrate a non-trivial understanding in simpler settings~\cite{garrido2025intuitive}. While multiple factors can explain this degradation in performance, a notable one is the stricter memory requirements. As illustrated in Figure~\ref{fig:comp-bench}, IntPhys 2 poses stricter requirements on short term memory than existing datasets, a property that video models can struggle with.

\section{Related Work}
\label{section:related}

\paragraph{Intuitive physics understanding benchmarking.}
Benchmarking intuitive physics understanding has been a focus of various datasets and challenges. \textsc{IntPhys} \citep{riochet2018intphys} tests machine perception of core physical phenomena through videos with possible and impossible physical events, inspired by developmental psychology experiments with infants. \textsc{GRASP} \citep{jassim_grasp_2024} provides a comprehensive evaluation framework for embodied AI that includes intuitive physics as one of its core domains. \textsc{InfLevel} \citep{weihs2022benchmarking} focuses on measuring intuitive physics understanding in simulation as well as in the real world. Other notable benchmarks include \textsc{PHYRE} \citep{bakhtin2019phyre}, which challenges models to solve physics puzzles through counterfactual reasoning, and \textsc{OOPS} \citep{epstein-etal-2020-oops}, which challenges models to predict when unintended physical actions occur in in-the-wild videos.

\paragraph{Broader physics understanding benchmarking.}
To evaluate broader physical understanding, benchmarks have been developed to test complex physical phenomena, including rigid bodies, fluids, soft bodies, and their interactions. \textsc{Physion} \citep{bear2021physion} presents a comprehensive evaluation suite that assesses visual physical prediction based on object properties in a scene. \textsc{CATER} \citep{girdhar2020cater} focuses on tracking and reasoning about moving objects, while \textsc{CLEVRER} \citep{yi2020clevrer} targets causal reasoning in video through descriptive, explanatory, predictive, and counterfactual questions. The Physical Interaction QA (\textsc{PIQA}) benchmark \citep{bisk2020piqa} tests physical commonsense knowledge through everyday human interactions. More recently, \textsc{Physion++} \citep{tung2023physion++} extends the original Physion benchmark to include more complex scenarios, and \textsc{Physics IQ} \citep{motamed2025physics} proposes a real-video benchmark to assess understanding of fundamental physical principles, including fluid dynamics, optics, solid mechanics, magnetism, and thermodynamics. More recently, \citet{wang2024compositional} introduced a new synthetic benchmark for assessing the following properties velocity, acceleration, and collisions.

\paragraph{Methods tackling physical understanding.}
Various computational approaches have been developed to tackle physical understanding challenges. World models \citep{ha2018worldmodels, hafner2019planet} learn latent dynamics of environments to predict future states and plan actions. Generative models, particularly those based on graph neural networks \citep{sanchez2020graph}, have shown promise in modeling physical dynamics by representing objects and their interactions. Joint Embedding Predictive Architectures (\textsc{JEPA}) \citep{lecun2022jepa} represent a self-supervised approach that learns to predict representations rather than raw observations. Large Language Models (\textsc{LLMs}) \citep{brown2020gpt3, openai2023gpt4} have demonstrated surprising capabilities in physical reasoning despite lacking explicit physical grounding. Hybrid approaches combining simulation-based reasoning with neural networks \citep{li2020visual} have also shown promise in physical understanding tasks by leveraging both data-driven learning and explicit physical knowledge.
More specialized methods have also been developed to tackle intuitive physics understanding, often relying on hardwired priors, trough the use of segmentation masks~\cite{piloto_intuitive_2022,riochet2020occlusion} or de-rendering~\cite{smith_adept_2019} for example.

\paragraph{Datasets generated with Unreal Engine.}
Unreal Engine has been widely adopted in the creation of synthetic datasets and benchmarks due to its advanced rendering capabilities and flexibility in simulating complex environments. For example, the CARLA simulator leverages Unreal Engine to provide a comprehensive autonomous driving benchmark, offering diverse driving scenarios and environmental conditions that are crucial for advancing research in autonomous vehicle perception and control \citep{Dosovitskiy2017}. Similarly, the AI2-THOR framework uses Unreal Engine to generate interactive environments for training and evaluating embodied AI agents, facilitating research in robotic manipulation and navigation \citep{Kolve2017}. UnrealCV also integrates with Unreal Engine to produce photorealistic images with ground truth annotations, supporting the development and evaluation of computer vision algorithms \citep{Qiu2017}. More recently, \citet{bordes2023pug} have used Unreal Engine for probing robustness of vision models. These projects highlight the engine's utility in generating high-quality datasets that enable researchers to explore new frontiers in AI and machine learning.

\section{Conclusion}
\label{section:discussion}

Our evaluations on IntPhys 2 reveal significant limitations in current models' intuitive physics reasoning capabilities, even for those that have shown promise in other benchmarks. The increased complexity and diversity of IntPhys 2, which mirrors real-world scenarios, exposes the models' inability to effectively process longer sequences, higher framerates, and utilize short-term memory. These limitations are evident in the almost-random performance of models on IntPhys 2, with only recent Multimodal Large Language Models like Gemini 2.5 Flash achieving non-trivial performance. Ultimately, IntPhys 2 highlights the need for novel model architectures and training methodologies that can bridge the gap between artificial and human cognition. By addressing the limitations of current models and the benchmark itself, we can pave the way for more robust and human-like AI systems that can approximate human intuitive physics understanding.
\paragraph{Limitations} While IntPhys 2 represents a significant advancement in benchmarking intuitive physics understanding, it also has limitations. Its reliance on synthetic environments may not fully capture the complexity of real-world physics, and the scope of physical principles covered is limited. Future research should focus on integrating real-world data, expanding the benchmark to include additional dimensions, and exploring more dynamic and interactive environments. Additionally, our evaluation setting has limitations, including the number of video frames that models can process, which differs from human perception. Humans can process full videos and retain long-term context, whereas current models are limited to sub-sampling a specific number of frames and cannot process multiple videos in a single context. These limitations highlight opportunities for future research and development of more advanced model architectures.

\clearpage
\newpage

\bibliography{paper}
\bibliographystyle{abbrvnat}

 \clearpage
 \newpage
 \beginappendix

\section{IntPhys2 Benchmark}

\begin{figure}
    \centering
    \includegraphics[width=0.8\linewidth]{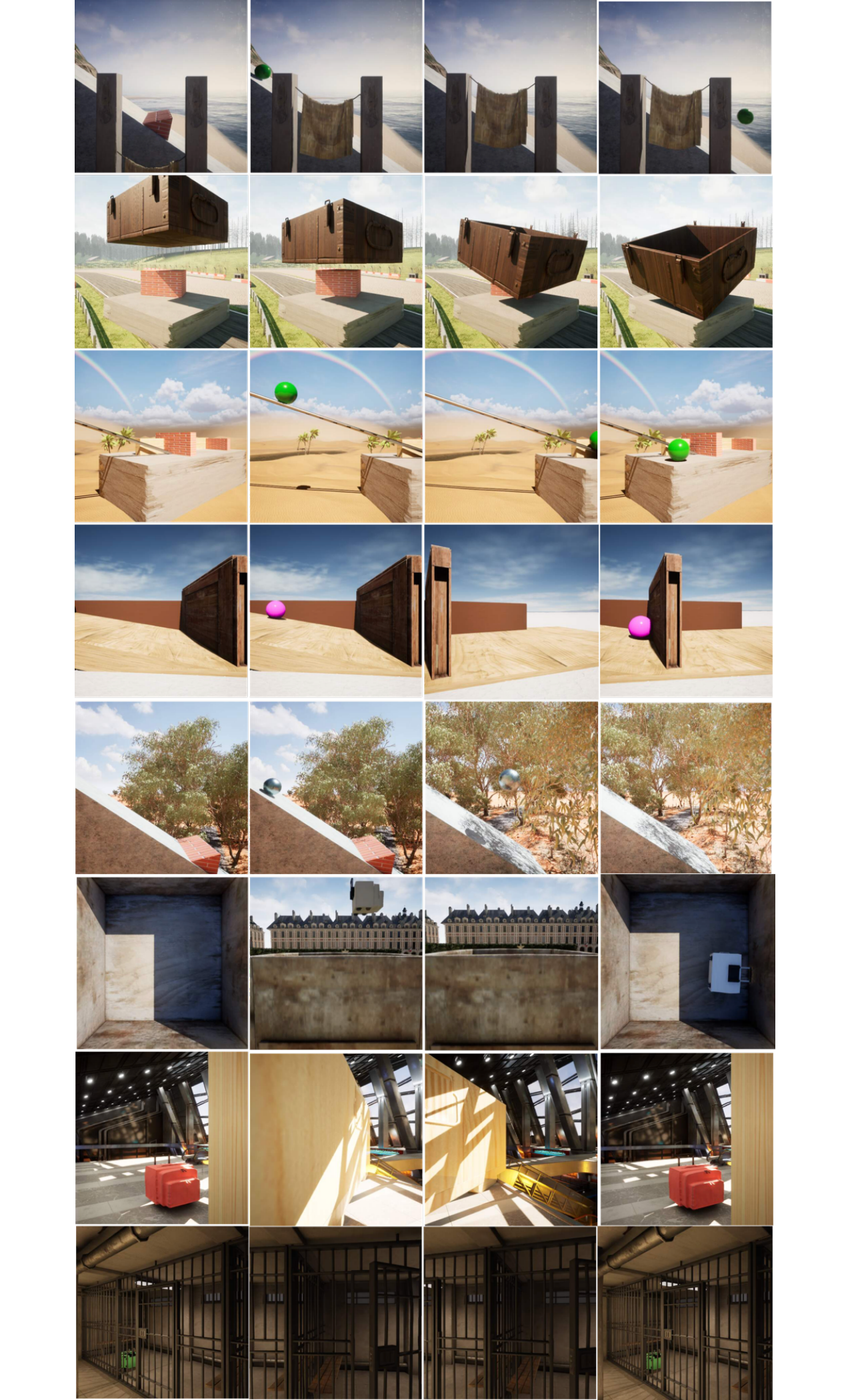}
    \caption{Example of different tasks and environments that are in the IntPhys2 benchmark.}
    \label{fig:Addition_example}
\end{figure}

\subsection{Designing the benchmark}
To design IntPhys2, we started by analyzing the limitations of the original IntPhys benchmark. One key issue was the use of simplistic wall-based occlusions, which reduced human performance due to their artificial nature. We aimed to create more realistic occlusion scenarios while maintaining the core principles of permanence, immutability, and spatio-temporal continuity.
Our first task involved a box occlusion paradigm where the camera observes an empty container, moves, and returns after an object is introduced. Plausible cases retain the object with consistent texture and shape, while implausible versions alter these properties. This tests object persistence under occlusion and sensitivity to material characteristics.
We also implemented a wall occlusion task where the camera moves behind a barrier and returns to its original position, evaluating object permanence during viewpoint changes. Additionally, we created ball dynamics tasks, including a slope task where a ball's trajectory changes predictably when encountering solid obstacles, and a rail task with constrained motion and occluders.
To expand our evaluation, we added a solidity condition with scenarios where movable boxes block rolling balls or alter occluder visibility. These tasks require agents to reason about physical interactions between objects. All tasks underwent rigorous validation to ensure clear physical principle violations in implausible cases and alignment with human plausibility judgments. We showcased some of the task in Figure \ref{fig:Addition_example}. 

\begin{figure}[ht]
    \centering
    \includegraphics[width=1.0\linewidth]{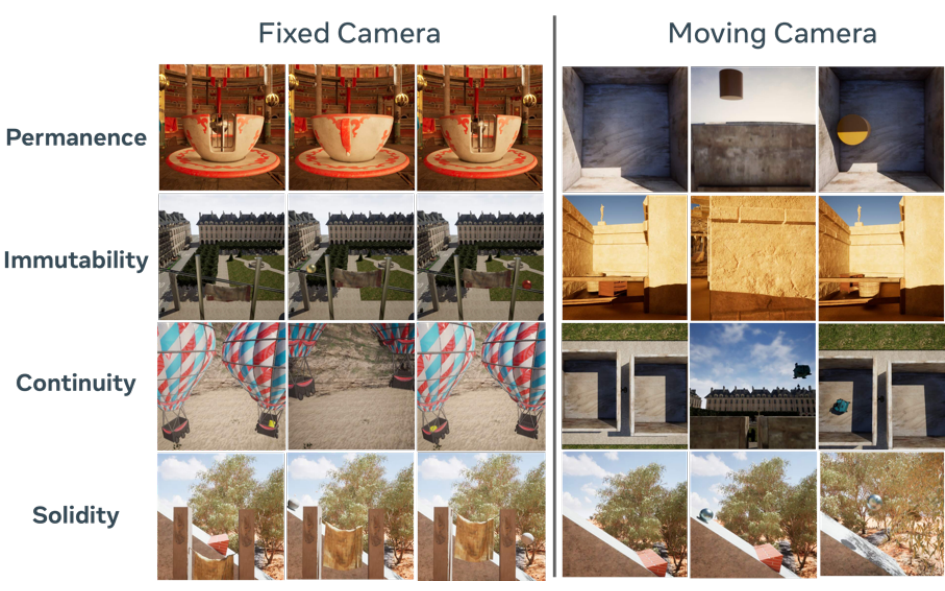}
    \caption{This example showcases the diversity of scenes and tasks in IntPhys 2. For each of the four conditions—Permanence, Immutability, Continuity, and Solidity—we feature scenes with either a static camera (similar to IntPhys1) or a moving camera, which offers a more natural perspective. Additionally, there is significant diversity in the environments, and the photorealistic capabilities of Unreal Engine enable us to enhance visual quality with realistic shadows and lighting.}
    \label{fig:condition_camera}
\end{figure}

In addition to camera's movements, we maintained a significant number of tasks with a fixed camera and moving occluder. Figure \ref{fig:condition_camera} shows video frame examples for each of the 4 conditions in IntPhys2, along with the two types of camera movement: fixed and moving. For fixed camera tasks, we used either a rotating cup to create occlusion or a hot air balloon that moved vertically. In other fixed camera tasks, we used a simple fabric sheet that moved upward to create occlusion. While this setup may not be highly realistic, it is more plausible than the wall-based occlusions used in the original IntPhys benchmark \cite{riochet2018intphys}.

A key design choice was the number of videos to generate. While simulated environments allow infinite data generation, this doesn't inherently increase diversity. Since tasks were manually designed, we could only make idiosyncratic changes to assets rather than altering entire scene structures. We currently lack tools to automatically generate diverse physical tests without human intervention. Instead, we intentionally limited the dataset size to: 1) avoid repetitive variations of the same scene, and 2) prevent training data contamination and overfitting that could result from excessive data points.

\subsection{Building the benchmark}
We built IntPhys2 using Unreal Engine\citep{unreal_disclaimer}, leveraging assets from the Unreal Marketplace and Fab to create our various tasks. A complete list of these assets is available on our project's GitHub repository. During runtime, we modified asset properties such as position, rotation, texture, and physical characteristics to increase scenario diversity. We also changed scene environments to test different physical configurations. Each task sequence was rendered as individual image frames, which were later compiled into videos using ffmpeg. To accelerate video production, we parallelized rendering across multiple V100 GPUs, completing the entire dataset generation in just 1 hour.

\subsection{Splits}
A key consideration in benchmark design is preventing training data contamination, which could bias model evaluations. While acknowledging this risk, we opted to release most of our video data along with metadata in our Main set. This decision was guided by our goal to create a diagnostic tool for assessing intuitive physics capabilities through zero-shot evaluation. The Main set enables fine-grained analysis of model strengths and weaknesses across different physical reasoning factors, making it our primary contribution. Metadata in this set should only be used for detailed error analysis, not for training. For researchers needing to validate basic scene understanding or adjust model hyperparameters, we provide a Debug set containing just 5 scenes. This limited size intentionally prevents generalization to the Main set unless a model possesses robust physical reasoning. Finally, we include a small Held-out set to detect potential contamination of our Main set in external training data. This layered approach balances diagnostic utility with contamination risks, maintaining the benchmark's integrity while supporting different evaluation needs.

\subsubsection{Debug Set}
\begin{figure}[ht]
    \centering
    \includegraphics[width=1.0\linewidth]{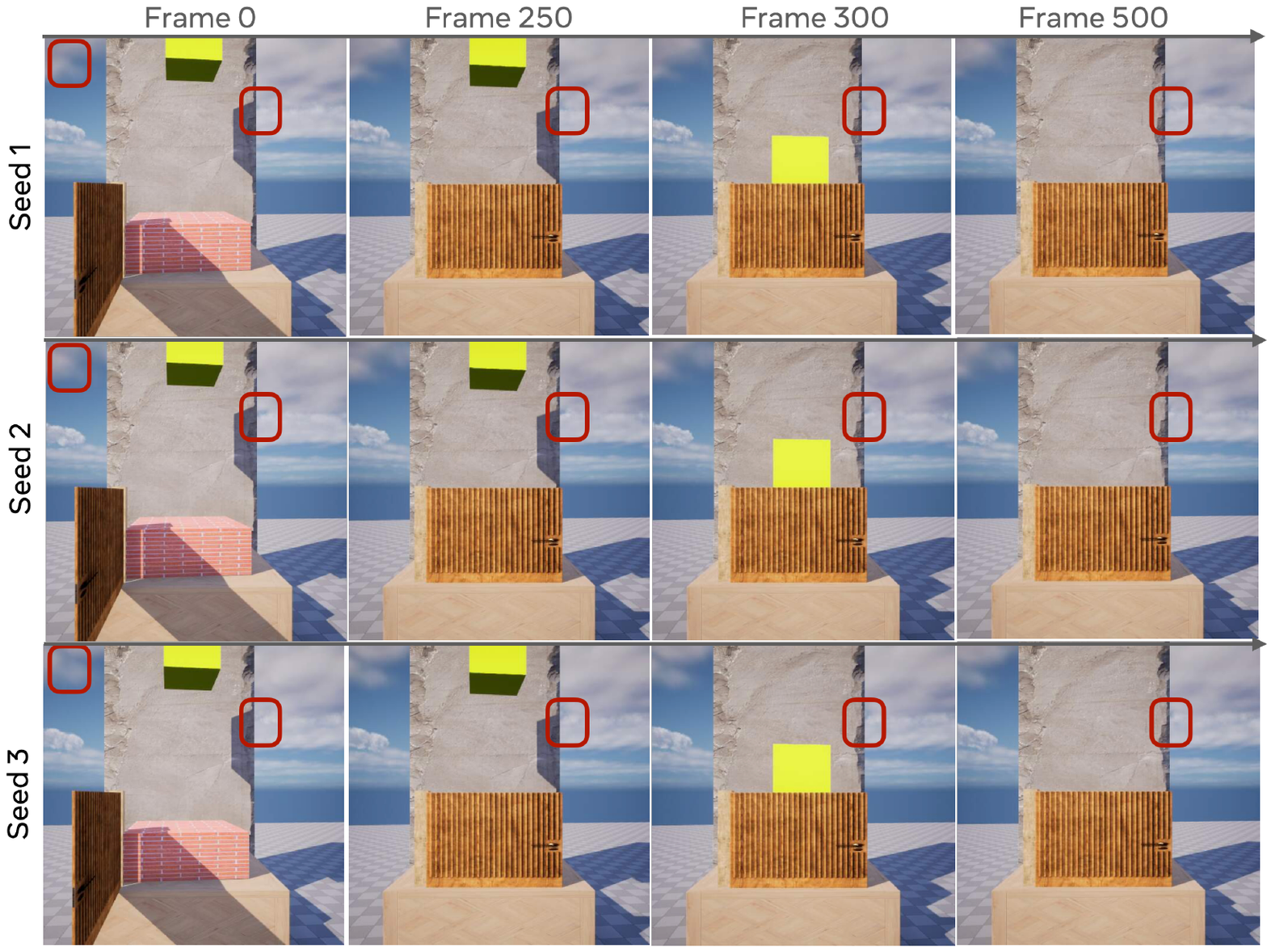}
    \caption{Example of a video from the Debug Set. This scene was sampled three different times and as we can see in the red circle, there are some slight variation in the sky that an human would probably not catch but that can have a significant effect on the model prediction as shown in Figure \ref{fig:NumberOfFrames}.}
    \label{fig:debug_set}
\end{figure}
The Debug set is designed to evaluate whether model predictions remain consistent across variations of the same scene. We created 60 videos using 5 different scenes, with 3 random seeds per video and 4 videos per scene. Figure \ref{fig:debug_set} illustrates the types of subtle visual variations that can occur between different videos of the same scene. For example, the circled red areas show differences in cloud appearance that might be imperceptible to humans but could potentially influence model predictions. Ideally, models should maintain consistent predictions despite these minor artifacts, demonstrating robustness to irrelevant visual variations. The tasks in the debug set are also designed to be extremely simple using bright assets that are easily visible.

\subsubsection{Main Set}
The Main set contains 1012 videos divided into three difficulty subsets. The Easy subset features simple tasks with basic environments - a tiled floor, sky background, and simple geometric assets like spheres and cubes in bright colors. This provides a baseline for expected model performance. The Medium subset introduces more complexity with varied environments and colored assets, increasing the challenge. The Hard subset is the most demanding, featuring rich backgrounds with realistic objects like air conditioning units and suitcases. Some scenes in this subset also include additional occlusions, such as moving leaves in front of the camera, further complicating the visual input.
Table \ref{tab:main_set_list} outlines the tasks and environments used in the Main set, demonstrating how we systematically increased difficulty while maintaining diagnostic utility. This tiered structure allows researchers to evaluate model performance across different levels of scene complexity and identify specific failure modes in physical reasoning.

\begin{table}[ht]
\centering
\tiny
\caption{Summary of video Metadata for the Main set}
\label{tab:video_metadata}
\begin{tabular}{|p{1.1cm}|p{1.9cm}|p{0.7cm}|c|p{5.5cm}|c|}
\hline
Condition & Task & Difficulty & Camera & Environment & Nb videos\\
\hline
\hline
Permanence & Box & \easy{Easy}, \medium{Medium}, \hard{Hard} & Moving & \easy{BasicLevel}, \medium{SaltFlats}, \medium{DesertMap}, \medium{RaceTrack}, \medium{TropicalIsland}, \hard{PLVDaylight}, \hard{Egypt}, \hard{RuralAustralia03}, \hard{ParkingGarage} & 36 \\ \hline
Permanence & MovingAroundOccluder & \easy{Easy}, \medium{Medium}, \hard{Hard} & Moving & \easy{BasicLevel}, \medium{SaltFlats}, \medium{DesertMap}, \medium{RaceTrack}, \medium{TropicalIsland}, \hard{RuralAustralia03}, \hard{ParkingGarage} & 32 \\ \hline
Permanence & JailStone & \easy{Easy}, \medium{Medium}, \hard{Hard} & Moving & \easy{BasicLevel}, \medium{SaltFlats}, \medium{DesertMap}, \medium{RaceTrack}, \medium{TropicalIsland}, \hard{Egypt}, \hard{RuralAustralia03}, \hard{ParkingGarage} & 44 \\ \hline
Permanence & RotatingCup & \easy{Easy}, \medium{Medium}, \hard{Hard} & Fixed & \easy{BasicLevel}, \medium{SaltFlats}, \medium{DesertMap}, \medium{RaceTrack}, \medium{TropicalIsland}, \hard{RuralAustralia03}, \hard{ParkingGarage} & 40 \\ \hline
Permanence & HotAirBallon & \easy{Easy}, \medium{Medium}, \hard{Hard} & Fixed & \easy{BasicLevel}, \medium{SaltFlats}, \medium{DesertMap}, \medium{RaceTrack}, \medium{TropicalIsland}, \hard{PLVDaylight}, \hard{Egypt}, \hard{RuralAustralia03}, \hard{ParkingGarage} & 36 \\ \hline
Permanence & SphereFallingDown & \easy{Easy}, \medium{Medium}, \hard{Hard} & Fixed & \easy{BasicLevel}, \medium{SaltFlats}, \medium{DesertMap}, \medium{RaceTrack}, \hard{PLVDaylight}, \hard{RuralAustralia03}, \hard{ParkingGarage} & 28 \\ \hline
Permanence & PrisonCell & \hard{Hard} & Moving & PrisonCell & 12 \\ \hline
Permanence & Restaurant & \hard{Hard} & Moving & Restaurant & 12 \\ \hline
Immutability & Box & \easy{Easy}, \medium{Medium}, \hard{Hard} & Moving & \easy{BasicLevel}, \medium{SaltFlats}, \medium{DesertMap}, \medium{RaceTrack}, \medium{TropicalIsland}, \hard{PLVDaylight}, \hard{Egypt}, \hard{RuralAustralia03}, \hard{ParkingGarage} & 36 \\ \hline
Immutability & MovingAroundOccluder & \easy{Easy}, \medium{Medium}, \hard{Hard} & Moving & \easy{BasicLevel}, \medium{SaltFlats}, \medium{DesertMap}, \medium{RaceTrack}, \medium{TropicalIsland}, \hard{RuralAustralia03}, \hard{ParkingGarage} & 32 \\ \hline
Immutability & JailStone & \easy{Easy}, \medium{Medium}, \hard{Hard} & Moving & \easy{BasicLevel}, \medium{SaltFlats}, \medium{DesertMap}, \medium{RaceTrack}, \medium{TropicalIsland}, \hard{Egypt}, \hard{RuralAustralia03}, \hard{ParkingGarage} & 44 \\ \hline
Immutability & RotatingCup & \easy{Easy}, \medium{Medium}, \hard{Hard} & Fixed & \easy{BasicLevel}, \medium{SaltFlats}, \medium{DesertMap}, \medium{RaceTrack}, \medium{TropicalIsland}, \hard{RuralAustralia03}, \hard{ParkingGarage} & 40 \\ \hline
Immutability & HotAirBallon & \easy{Easy}, \medium{Medium}, \hard{Hard} & Fixed & \easy{BasicLevel}, \medium{SaltFlats}, \medium{DesertMap}, \medium{RaceTrack}, \medium{TropicalIsland}, \hard{PLVDaylight}, \hard{Egypt}, \hard{RuralAustralia03}, \hard{ParkingGarage} & 36 \\ \hline
Immutability & SphereFallingDown & \easy{Easy}, \medium{Medium}, \hard{Hard} & Fixed & \easy{BasicLevel}, \medium{SaltFlats}, \medium{DesertMap}, \medium{RaceTrack}, \hard{PLVDaylight}, \hard{RuralAustralia03}, \hard{ParkingGarage} & 28 \\ \hline
Immutability & PrisonCell & \hard{Hard} & Moving & PrisonCell & 12 \\ \hline
Immutability & Restaurant & \hard{Hard} & Moving & Restaurant & 12 \\ \hline
Continuity & Box & \easy{Easy}, \medium{Medium}, \hard{Hard} & Moving & \easy{BasicLevel}, \medium{SaltFlats}, \medium{DesertMap}, \medium{RaceTrack}, \medium{TropicalIsland}, \hard{PLVDaylight}, \hard{Egypt}, \hard{RuralAustralia03}, \hard{ParkingGarage} & 36 \\ \hline
Continuity & MovingAroundOccluder & \easy{Easy}, \medium{Medium}, \hard{Hard} & Moving & \easy{BasicLevel}, \medium{SaltFlats}, \medium{DesertMap}, \medium{RaceTrack}, \medium{TropicalIsland}, \hard{RuralAustralia03}, \hard{ParkingGarage} & 32 \\ \hline
Continuity & JailStone & \easy{Easy}, \medium{Medium}, \hard{Hard} & Moving & \easy{BasicLevel}, \medium{SaltFlats}, \medium{DesertMap}, \medium{RaceTrack}, \medium{TropicalIsland}, \hard{Egypt}, \hard{RuralAustralia03}, \hard{ParkingGarage} & 44 \\ \hline
Continuity & RotatingCup & \easy{Easy}, \medium{Medium}, \hard{Hard} & Fixed & \easy{BasicLevel}, \medium{SaltFlats}, \medium{DesertMap}, \medium{RaceTrack}, \medium{TropicalIsland}, \hard{RuralAustralia03}, \hard{ParkingGarage} & 40 \\ \hline
Continuity & HotAirBallon & \easy{Easy}, \medium{Medium}, \hard{Hard} & Fixed & \easy{BasicLevel}, \medium{SaltFlats}, \medium{DesertMap}, \medium{RaceTrack}, \medium{TropicalIsland}, \hard{PLVDaylight}, \hard{Egypt}, \hard{RuralAustralia03}, \hard{ParkingGarage} & 36 \\ \hline
Continuity & SphereFallingDown & \easy{Easy}, \medium{Medium}, \hard{Hard} & Fixed & \easy{BasicLevel}, \medium{SaltFlats}, \medium{DesertMap}, \medium{RaceTrack}, \hard{PLVDaylight}, \hard{RuralAustralia03}, \hard{ParkingGarage} & 28 \\ \hline
Continuity & PrisonCell & \hard{Hard} & Moving & PrisonCell & 12 \\ \hline
Continuity & Restaurant & \hard{Hard} & Moving & Restaurant & 12 \\ \hline
Solidity & FixedJumpSolidity & \easy{Easy}, \medium{Medium}, \hard{Hard} & Fixed & \easy{BasicLevel}, \medium{SaltFlats}, \medium{DesertMap}, \medium{RaceTrack}, \medium{TropicalIsland}, \hard{PLVDaylight}, \hard{RuralAustralia03}, \hard{ParkingGarage} & 32 \\ \hline
Solidity & SolidityFallingFlat & \easy{Easy}, \medium{Medium}, \hard{Hard} & Fixed & \easy{BasicLevel}, \medium{SaltFlats}, \medium{DesertMap}, \medium{RaceTrack}, \medium{TropicalIsland}, \hard{PLVDaylight}, \hard{Egypt}, \hard{RuralAustralia03}, \hard{ParkingGarage} & 36 \\ \hline
Solidity & SphereFallingDownSoldity & \easy{Easy}, \medium{Medium}, \hard{Hard} & Moving & \easy{BasicLevel}, \medium{SaltFlats}, \medium{DesertMap}, \medium{RaceTrack}, \hard{PLVDaylight}, \hard{RuralAustralia03}, \hard{ParkingGarage} & 28 \\ \hline
Solidity & FixedMarryPoppins & \easy{Easy}, \medium{Medium}, \hard{Hard} & Fixed & \easy{BasicLevel}, \medium{SaltFlats}, \medium{DesertMap}, \medium{RaceTrack}, \medium{TropicalIsland}, \hard{PLVDaylight}, \hard{Egypt}, \hard{RuralAustralia03}, \hard{ParkingGarage} & 36 \\ \hline
Solidity & BoxSoldity & \easy{Easy}, \medium{Medium}, \hard{Hard} & Moving & \easy{BasicLevel}, \medium{SaltFlats}, \medium{DesertMap}, \medium{RaceTrack}, \medium{TropicalIsland}, \hard{PLVDaylight}, \hard{Egypt}, \hard{RuralAustralia03}, \hard{ParkingGarage} & 40 \\ \hline
Solidity & Scaffoling & \easy{Easy}, \medium{Medium}, \hard{Hard} & Moving & \easy{BasicLevel}, \medium{SaltFlats}, \medium{DesertMap}, \medium{RaceTrack}, \medium{TropicalIsland}, \hard{PLVDaylight}, \hard{Egypt}, \hard{RuralAustralia03}, \hard{ParkingGarage} & 36 \\ \hline
Solidity & CameraSolidity & \easy{Easy}, \medium{Medium}, \hard{Hard} & Moving & \easy{BasicLevel}, \medium{SaltFlats}, \medium{DesertMap}, \medium{RaceTrack}, \medium{TropicalIsland}, \hard{RuralAustralia03}, \hard{ParkingGarage} & 40 \\ \hline
Solidity & JumpSolidity & \easy{Easy}, \medium{Medium}, \hard{Hard} & Moving & \easy{BasicLevel}, \medium{SaltFlats}, \medium{DesertMap}, \medium{RaceTrack}, \medium{TropicalIsland}, \hard{PLVDaylight}, \hard{RuralAustralia03}, \hard{ParkingGarage} & 44 \\ \hline
\end{tabular}
\label{tab:main_set_list}
\end{table}

\subsubsection{Held-Out Set}
The Held-out set contains 344 videos sampled from the same distribution as the Hard subset of the Main set. We do not provide metadata for this set to prevent potential training data contamination. Researchers can evaluate their models on this set through a leaderboard hosted on HuggingFace, ensuring a standardized comparison of physical reasoning capabilities.

\subsection{DataSheet}
Here is the full datasheet for the IntPhys2 benchmark.

\begin{longtable}{|p{6cm}|p{8cm}|}
    \toprule
    \multicolumn{2}{c}{\textsc{\textbf{Motivation}}} \\
    \midrule
    \textbf{For what purpose was the dataset created?} & IntPhys2 was created to assess wether deep learning model have similar intuitive physic capabilities as human. Simiarly to IntPhy1, models are showns different videos in which some of them represent something that is physicall plausibler while other are not.\\ \midrule
    \textbf{Who created the dataset and on behalf of which entity?} & This dataset was created by the FAIR team at Meta. \\ \midrule
    \textbf{Who funded the creation of the dataset?} & Meta. \\ \midrule
    \multicolumn{2}{c}{\textsc{\textbf{Composition}}} \\ \midrule
    \textbf{What do the instances that comprise the dataset represent?} &  The instances represent videos of different shapes and objects in a diverse set of background.\\ \midrule
    \textbf{How many instances are there in total?} & 1416 videos. \\ \midrule
    \textbf{Does the dataset contain all possible instances or is it a sample (not necessarily random) of instances from a larger set?} & Each scene in the dataset is represent by a video quadruplet that reprent two event that are possible while two other are not. The videos are matched in such a was as the possible even in a scenario becomes impossible when there is a slight change in the scene. \\ \midrule
    \textbf{Is there a label or target associated with each instance?} & Yes, a csv file. Each instance have a row in this csv files with all the factors of variation used to generate this image. The csv files contains the following columns:\newline\newline
\quote{SceneIndex,name,filename,gamename, condition,env,type,occluder,Difficulty,Camera}\newline\newline
    
    \\ \midrule
    \textbf{Is any information missing from individual instances?} & Only the metadata in the held out set are not included.    \\ \midrule
    \textbf{Are relationships between individual instances made explicit?} & Yes, each of the video belonging to a given quadruplet share the same scene index. \\ \midrule
    \textbf{Are there recommended data splits?} & There are 3 different splits of the data. The first one, the debug set, can be use to calibrate the model, verify the number of frame required and check that the performances are not random. This is a very easy subset with always the same background, fixed camera and in which we sample the same scene multiple time to assess wether the model is deterministic or not, there is around 100 videos in this set. Then there is the main set, which consists of 3 subsplits: Easy, Medium, Hard which is the set for which most people will run their evaluations on. The last set is an held out set that is use to confirm that the performances gain are not due to any pre-training on the data. \\ \midrule
    \textbf{Are there any errors, sources of noise, or redundancies in the dataset?} & In the debug set, we sampled multiple time the same videos of the exact same scene to assess wether a model can be sensitive to any video or scene recording noise (like cloud moving in the background, or light that is slightly changing). \\ \midrule
    \textbf{Is the dataset self-contained, or does it link to or otherwise rely on external resources?} & The dataset is self-contained however the assets that were used to build the dataset belongs to external sources which are listed in the github at \url{https://github.com/facebookresearch/IntPhys2}.    \\ \midrule
    \textbf{Does the dataset contain data that might be considered confidential?} & No. \\ \midrule
    \textbf{Does the dataset contain data that, if viewed directly, might be offensive, insulting, threatening, or might otherwise cause anxiety?} & No. \\ \midrule
    \multicolumn{2}{c}{\textsc{\textbf{Collection}}} \\ \midrule
    \textbf{How was the data associated with each instance acquired?} & The data (3D assets) were acquired through the Unreal Engine Marketplace \url{https://www.unrealengine.com/marketplace/en-US/store} and Fab \url{https://fab.com}. Assets were then incorporated into the Unreal Engine to generate realistic 3D scenes and corresponding images. The 3D assets were manually selected to ensure high quality. \\ \midrule
    \textbf{What mechanisms or procedures were used to collect the data?} & Manual human curation. Assets were manually collected.    \\ \midrule
    \textbf{If the dataset is a sample from a larger set, what was the sampling strategy?} & There is an infinite amount of data that could have been sampled, we mostly wanted to maximize dievrsity of the task and scenes to provide a challenging benchmark for model that is yet easy to solve by humans,   \\ \midrule
    \textbf{Who was involved in the data collection process and how were they compensated?} & Only the authors of this work were involved in crafting the dataset.   \\ \midrule
    \textbf{Over what timeframe was the data collected?} & The assets were bought between November 2024 and February 6, 2025 \\ \midrule
    \textbf{Were any ethical review processes conducted?} & No. \\ \midrule
    \textbf{Did you collect the data from the individuals in question directly, or obtain it via third parties or other sources (e.g., websites)?} & Third parties: Unreal Engine Marketplace \url{https://www.unrealengine.com/marketplace/en-US/store} and fab \url{https://fab.com/}.   \\ \midrule
    \textbf{Were the individuals in question notified about the data collection? If so, please describe (or show with screenshots or other information) how notice was provided, and provide a link or other access point to, or otherwise reproduce, the exact language of the notification itself.} & There is no personally identifiable information in our datasets as they are purely synthetic and contain no images of people. We purchased 3D assets from different marketplaces where required, however we did not explicitly contact the individual creators. \\ \midrule
    \textbf{Did the individuals in question consent to the collection and use of their data? If so, please describe (or show with screenshots or other information) how consent was requested and provided, and provide a link or other access point to, or otherwise reproduce, the exact language to which the individuals consented.} & N/A. See above. \\ \midrule
    \textbf{If consent was obtained, were the consenting individuals provided with a mechanism to revoke their consent in the future or for certain uses? If so, please provide a description, as well as a link or other access point to the mechanism (if appropriate).} & N/A. See above. \\ \midrule
    \textbf{Has an analysis of the potential impact of the dataset and its use on data subjects (e.g., a data protection impact analysis) been conducted? If so, please provide a description of this analysis, including the outcomes, as well as a link or other access point to any supporting documentation.} & No data about specific individuals is included in these data. See above. \\ \midrule
    \multicolumn{2}{c}{\textsc{\textbf{Preprocessing}}} \\ \midrule
    \textbf{Was any preprocessing/cleaning/labeling of the data done?} & N/A.   \\ \midrule
    \textbf{Was the “raw” data saved in addition to the preprocessed/cleaned/labeled data?} & N/A. \\ \midrule
    \textbf{ Is the software that was used to preprocess/clean/label the data available?} & N/A. \\ \midrule
    \multicolumn{2}{c}{\textsc{\textbf{Uses}}} \\ \midrule
    \textbf{Has the dataset been used for any tasks already?} & Yes, these data were used for the experiments that were presented in this paper.    \\ \midrule
    \textbf{Is there a repository that links to any or all papers or systems that use the dataset?} & No. \\ \midrule
    \textbf{What (other) tasks could the dataset be used for?} & No other tasks, only for model evaluation.\\ \midrule
    \textbf{Is there anything about the composition of the dataset or the way it was collected and preprocessed/cleaned/labeled that might impact future uses?} & No. \\ \midrule
    \textbf{Are there tasks for which the dataset should not be used?} & These datasets \textbf{should not be used} for generative modeling purposes. \\ \midrule
    \multicolumn{2}{c}{\textsc{\textbf{Distribution}}} \\ \midrule
    \textbf{Will the dataset be distributed to third parties outside of the entity on behalf of which the dataset was created?} & Yes, the dataset will be publicly distributed. \\ \midrule
    \textbf{How will the dataset will be distributed?} & Tarball on a website. \\ \midrule
    \textbf{Will the dataset be distributed under a copyright or other intellectual property (IP) license, and/or under applicable terms of use (ToU)?} & The license of the dataset is \textbf{cc-by-nc with the mention that these data should not be used for generative AI purposes}. \\ \midrule
    \textbf{Have any third parties imposed IP-based or other restrictions on the data associated with the instances?} & See \citet{unreal_disclaimer}. \\ \midrule
    \textbf{Do any export controls or other regulatory restrictions apply to the dataset or to individual instances?} & N/A \\ \midrule
    \multicolumn{2}{c}{\textsc{\textbf{Maintenance}}} \\ \midrule
    \textbf{Who will be supporting/hosting/maintaining the dataset?} & Meta. \\ \midrule
    \textbf{How can the owner/curator/manager of the dataset be contacted?} & Please contact the corresponding author of this paper. \\ \midrule
    \textbf{Is there an erratum?} & No. \\ \midrule
    \textbf{Will the dataset be updated?} & Probably not. \\ \midrule
    \textbf{If the dataset relates to people, are there applicable limits on the retention of the data associated with the instances (e.g., were the individuals in question told that their data would be retained for a fixed period of time and then deleted)? If so, please describe these limits and explain how they will be enforced.} & N/A. \\ \midrule
    \textbf{Will older versions of the dataset continue to be supported/hosted/maintained? If so, please describe how. If not, please describe how its obsolescence will be communicated to dataset consumers.} & If any new version should be made available, it will be communicated through the Github and Hugging face repository. \\ \midrule
    \textbf{If others want to extend/augment/build on/contribute to the dataset, is there a mechanism for them to do so?} & No mechanisms are in place yet, but they can contact the authors of this paper if they would like to contribute. \\ \bottomrule
    \caption{\textbf{Datasheet for IntPhys2}, following the framework introduced by \citet{gebru2021datasheets}.}
    \label{tab:datasheet}
\end{longtable}

\section{Human evaluation}
\label{appendix:human_eval}

For human evaluation, we utilized the services of Moravia. Each annotator reviewed 96 videos and rated their plausibility on a Likert scale from 1 to 4. Each annotator have seen all the quadruplet for a given scene to ensure consistency with respect to human answers. Each video was evaluated by three annotators, and the consensus answer was determined by a majority vote.

\section{MLLMs evaluations}

For the MLLMs evaluations, we have used the prompts in Table \ref{tab:prompts}. It's not straightforward to find the most suited prompts, so we took inspiration from the work of \citet{garrido2025intuitive} who evaluated several MLLMs on the IntPhys benchmark. To assess the impact of the prompt on model's performances, we added two additional prompt in which we either ask the model to provide a yes/no answer or a binary digit answer. This variability of prompt allow us to asses how much a model can be robust to a different set of instructions. 

To perform the ablations over the different number of frames, we have just generated the videos using different frame rates. Since most API and library subsample the video a 1fps, we just ensure that the video was extended enough to have an enough number of frames.

For GPT-4o and Gemini, we have use the API while for Qwen-VL, we have used the transformers\citep{wolf2020huggingfacestransformersstateoftheartnatural} library. 

\begin{table}[h]
    \centering
    \small
    \begin{tabular}{>{\raggedright\arraybackslash}p{0.45\textwidth} >{\raggedright\arraybackslash}p{0.45\textwidth}}
        \toprule
        \textbf{Prompt 1} & \textbf{Prompt 2 and \textcolor{blue}{3}} \\
        \midrule
        
    The video you’re seeing was generated by a simulator. Given how objects behave on earth, is the behavior of the objects plausible? Your answer should be based on the events in the video and ignore the quality of the simulation. You should also ignore the framerate of the video, as it may be slowed down. Finish your answer by Yes or No.  &
    
    Analyze the physics simulations in the video. Determine if the video depicts object behavior that is consistent with Earth's physical laws. Ignore simulation quality and playback speed. CRITICAL OUTPUT INSTRUCTION: Your response must be (\textcolor{blue}{yes or no}) only a binary value indicating whether the object behavior is consistent with Earth's physical laws: (\textcolor{blue}{yes}) 1: Object behavior is consistent with Earth's physical laws. (\textcolor{blue}{no}) 0: Object behavior is not consistent with Earth's physical laws. Please provide your analysis in the required format. Do not include any other text, explanation, context, preamble, or punctuation before or after the answer.\\
    \bottomrule
    \end{tabular}
    \caption{Comparison of Two Different Prompts. The first prompt is shorted and more open-ended. While the second prompt is much more specific. It also comes with two version, one in which the models answer a binary value while the other answer a yes/no question.}
    \label{tab:prompts}
\end{table}

\clearpage
 \section{Prediction based evaluation \label{sec:pred-protocol}}
\subsection{Protocol}
\begin{figure}[!tbp]
    \centering
    \includegraphics[width=0.9\linewidth]{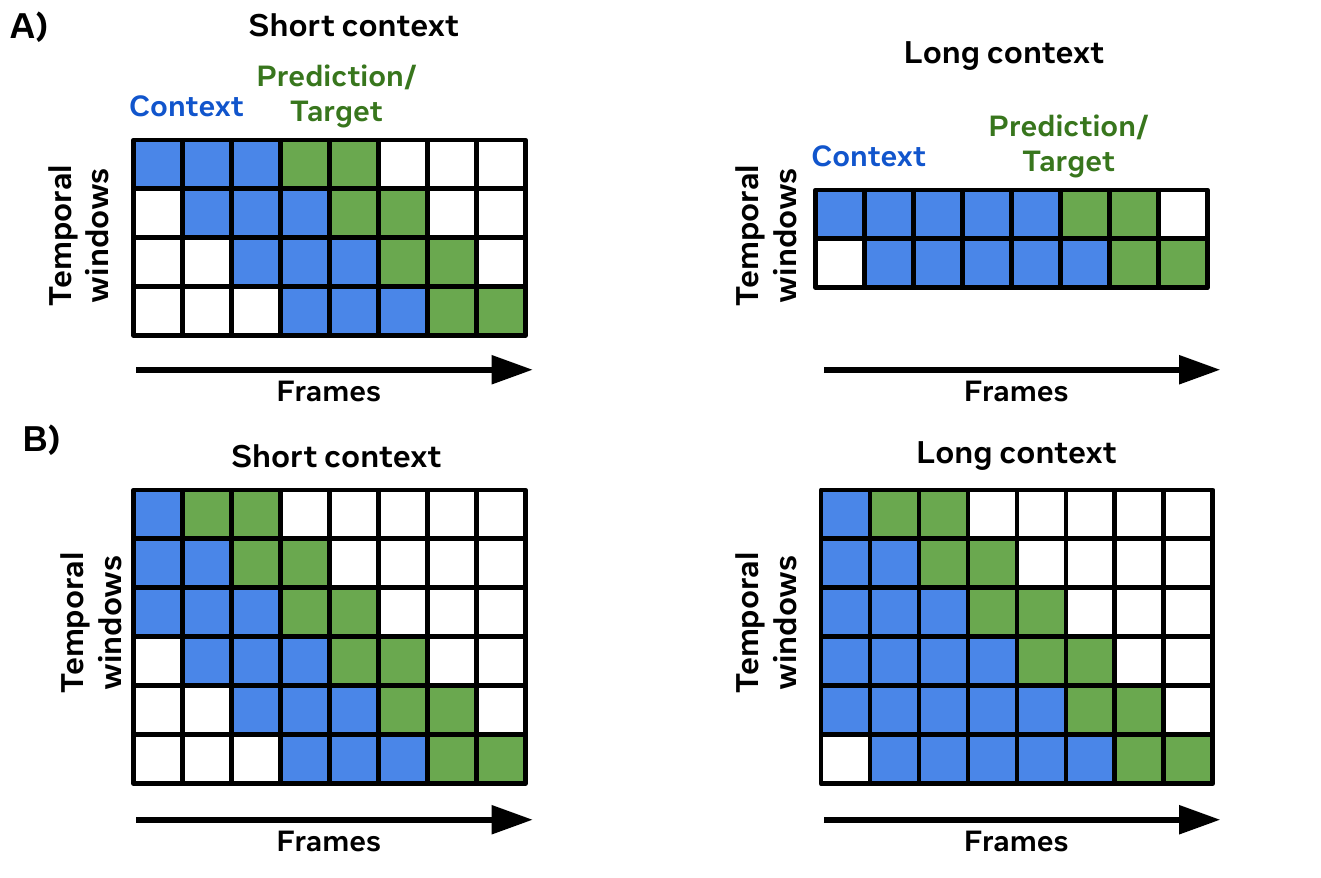}
    \caption{\textbf{Protocol for prediction based models.} Previous protocols are tailored for models with small contexts. When increasing the context size, the predictions start later in the video, possibly missing the impossible event \textbf{(A)}. We propose to consider the context as a maximum context, ensuring that the majority of frames are predicted, and that any context size can be used \textbf{(B)}.}
    \label{fig:pred-protocol}
\end{figure}

Our evaluation protocol is based on the one developed in~\cite{garrido2025intuitive}, which demonstrated non trivial performance on existing intuitive physics datasets.

Considering a video $V$ of $N$ frames and a model capable of handling $M < N$ frames, we break the video into overlapping sliding windows of $M$ frames. We can defined a stride $S$ between these sliding windows, giving windows of frames $0 \rightarrow M-1$, $S \rightarrow M-1+S$,..., $N-1-M \rightarrow N-1$. These successive windows correspond as the blue and green frames in figure~\ref{fig:pred-protocol}. 
A given window is then split into two parts: the context (used to make the prediction), and the target (what is predicted). We denote the length of the context by $C$, and the length of the prediction by $T = M-C$. We can then defined a per window surprise as 
\begin{equation}
    \texttt{Surprise}_w = d\left(p\left(f\left(V_{w:w+C}\right)\right), f'\left(V_{w+C:w+M}\right) \right),
\end{equation}
where $w$ is the starting frame of the window, $d$ a distance measure, $f$ the encoder used for the context, $f'$ the one for the target, and $p$ is the predictor, tasked with predicting the future. For example, in VideoMAEv2, $f$ is the encoder, $f'$ the identity function, and $p$ the decoder.

The surprise can then be aggregated over all windows to obtain a surprise measure over the whole video. Two common choices are the average and maximal surprise defined as 
\begin{equation}
    \texttt{AvgSurprise} =\frac{1}{\#W} \sum_{w \in W:=\{0,S,2S,...\}} \texttt{Surprise}_w \quad\text{and}\quad \texttt{MaxSurprise} =\max_{w\in W}\left(\texttt{Surprise}_w\right).
\end{equation}

The former is particularly meaningful for pairwise evaluations as it detects subtle differences between videos with a similar baseline surprise, and the latter is particularly meaningful for single video classification as it ignores the same baseline surprise. We use average surprise unless specified otherwise.

The main limitation of this protocol becomes apparent when considering a large window size $M \simeq N$, as illustrated on the right of figure~\ref{fig:pred-protocol}. In this setting the predictions start very late in the video, and we may miss the impossible event we want to detect. While most model have a small window size, usually not more than 32 frames, the increased memory requirements of IntPhys 2 lend it well to models with longer window sizes, and thus a default protocol should allow it. We thus propose to consider the window size $M$ as a maximal size, and also proceed with all predictions using smaller ones, by progressively growing the context size. This is illustrated at the bottom of figure~\ref{fig:pred-protocol}. This ensures that every frame (apart from the first one) is predicted no matter the window size.

To avoid the aforementioned constraint, the performance reported with the protocol of~\cite{garrido2025intuitive} is taken as the maximal one over different frame rates and context sizes, two hyperparameters that affect the memory of the model, as well as the first predicted frame. This has limitations when the number of hyperparameters tested increases, since this maximum operation increases performance in the presence of noise.

\paragraph{Special case of Cosmos.} While for VideoMAEv2 and V-JEPA the prediction must happend in pixel and representation space respectively, we can choose between both for Cosmos.
The autoregressive model used in Cosmos predicts latent codes, or rather their indices in a codebook. To predict in representation space we thus obtain the code associated with the predicted index, and to predict in pixel space, we feed the predicted tokens through the decoder. The target can then either be encoded for representation space prediction, or simply kept as the future pixel.

We find that on IntPhys, predicting in pixel space achieves random performance, but predicting in representation space yields a performance of around 85\%. We thus opt for the latter.

\subsection{Comparison with existing protocol}

As described in the previous section, the protocol from~\cite{garrido2025intuitive} demonstrated non trivial performance for V-JEPA and VideoMAEv2. One can thus wonder if the new protocol we use performs worse and could be the reason of the poor performance we observe.
We start by noting that all prediction windows with the previous protocol are present in ours, and that we are mainly extending the "surprise over time curve" by filling in its beginning. Furthermore, we evaluate VideoMAEv2 and V-JEPA with both protocols on IntPhys 2 and compare the performance in table~\ref{tab:old_vs_new_protocol}.

\begin{table}[]
    \centering
    \caption{\textbf{Comparison between evaluation protocols.} We find that the protocol from \citet{garrido2025intuitive} yields equivalent performance.}
    \begin{tabular}{lccc}
        \toprule
        Model & V-JEPA + RoPE & VideoMAEv2  \\
        \midrule
        Previous protocol~\cite{garrido2025intuitive} & 52.96  & 54.94 \\
        Our protocol & 53.75  & 53.75 \\
        \bottomrule
    \end{tabular}
   
    \label{tab:old_vs_new_protocol}
\end{table}

We find that models perform similarly with both protocols, though we believe that our protocol will scale better to longer context lengths, something current models struggle with. For example, in both cases the best performance is achieved with a window size of 16 and 6 as framerate (i.e. less than 3 seconds), which makes most samples from IntPhys 2 impossible.

\subsection{Hyperparameters~\label{sec:hyperparams-pred}}

\begin{table}[]
    \centering
    \caption{\textbf{Best hyperparamters for prediction models.} We report the best hyperparameters found across models as judged by the main set of IntPhys 2. A prediction length of -1 indicates to predict to the end of the window rather than a fixed number of frames.}
    \begin{tabular}{l c c c c}
        \toprule
         Model & Window size & Prediction length & Framerate & Other \\
         \midrule
         V-JEPA + RoPE& 16 & 2 & 6 & N/A   \\
         V-JEPA 2 & 48 & -1 & 6 & N/A   \\
         VideoMAEv2& 16 & -1 & 6 & N/A  \\
         Cosmos & 33 & 24 & 6  & temperature = 0, top\_p = 0.8 \\
         
         \bottomrule
    \end{tabular}
    \label{tab:my_label}
\end{table}

Different tasks require different amount of memory, and models may fare better when extending the window size, or reducing the frame rate. Similarly, models can often predict at a short temporal range, or at a longer one. We thus ablate these three parameters in our experiments.

For V-JEPA and V-JEPA 2, the use of RoPe makes it apt to use longer window sizes, as evidenced in~\cite{garrido2025intuitive}. We thus use window sizes of 16,32 or 48 frames with a fixed framerate of 6 fps.

For VideoMAEv2, the use of sincos embeddings makes it more suited to reducing the framerate, so we keep the window size fixed at 16 and vary the framerate at 2,3 and 6 fps.

For both methods we experimented in both directions (framerate and window size), but to have similar grid search sizes landed on these two axes as they yielded better signal.

Cosmos has larger constraints on the window size and prediction length, where official inference only supports 9 frames of context and 24 frames of prediction when using the autoregressive model without additional conditioning. Experiments with predicting fewer frames did not yield meaningful signal so we keep the parameters as is. This is also bottlenecked by the large inference cost of Cosmos, where the 4B model already takes around 40 times as long as V-JEPA or VideoMAE to run. Nonetheless, we ablate the temperature as 0 or 1, and top\_p as 0.7, 0.8 or 0.9.

When possible, we use different context length for a given scenario and report the maximal accuracy across them. We use 4,6,8,10,12 and 14 for VideoMAEv2, V-JEPA and V-JEPA 2, and stay with the 9 of Cosmos. While this does not make for a perfect comparison, the constraints of each models must be respected and we do not see any meaningful performance increase above chance level across any model, that could result from better hyperparameter tuning.

\subsection{Cost of experiments}

Evaluations were ran on NVIDIA H100 GPUs. For V-JEPA, V-JEPA 2 and VideoMAEv2 all experiments are ran on a compute node with 8 GPUs. On the main set, experiments take between 15 and 30 minutes depending on hyperparameters. For Cosmos, due to the size of the model, experiments are run on 8 nodes of 8 GPUs each, with experiments on the main set taking around 2.5 hours.

\clearpage
\section{Qualitative results for prediction based models on the debug set~\label{sec:qual-debug}}

The debug set is designed to help practitioners tune their method before evaluating on the full dataset. It provides simple videos and is adequately sized for qualitative analysis, such as studying surprise curves over time or visualizing model predictions. We look at these aspects in this section, both to provide a qualitative analysis of the previously presented results, and to showcase how this debug set can be used.

For both subsections that follow, we use the same video from the debug set for ease of comparison.

\subsection{V-JEPA surprise visualization}

\begin{figure}[tbp]
    \centering
    \includegraphics[width=\linewidth]{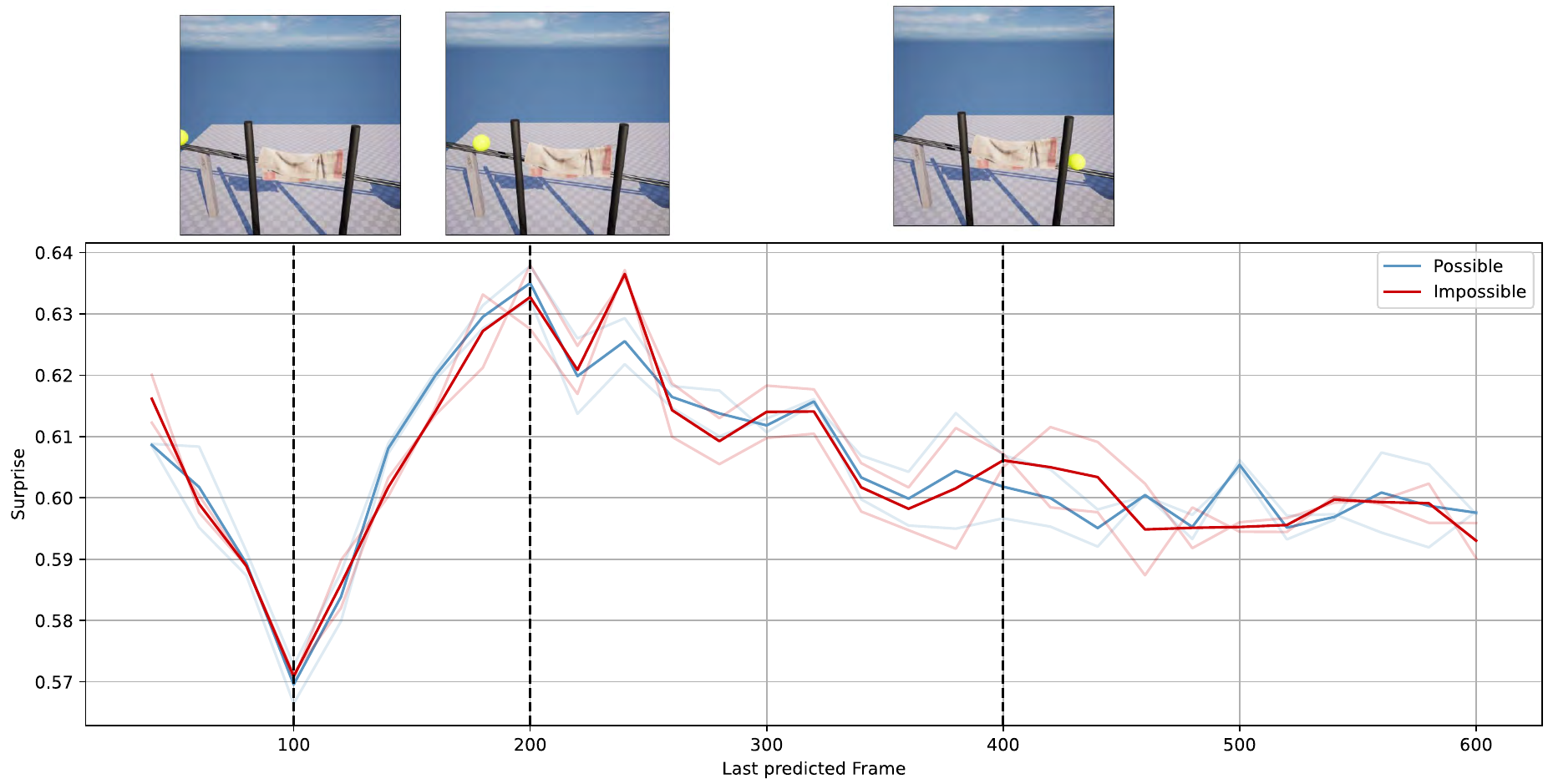}
    \caption{\textbf{Surprise visualization with V-JEPA.} We find that V-JEPA exhibits high surprise when there is movement in the video, but there is no constant increase of surprise when the object is supposed to reappear. We plot the surprises for all possible and impossible videos in this scene, with the average surprise across possible and impossible video being highlighted.}
    \label{fig:quali_jepa}
\end{figure}

Taking a scene from the debug set, we compute the surprise of a V-JEPA model using a maximal context length of 20 frames, tasked with predicting the next two, using a frame skip of 10. We note that changing the number of context frames or prediction length did not yield different behaviour for the experiments to follow.

We then visualize the surprise mapped to the last predicted frame in figure~\ref{fig:quali_jepa}. As we are considering the same scene across different seeds, we plot the individual surprise curves, as well as the average one in order to see if any trend appears. We find a decreasing surprise in the beginning of the video, when only the occluder  is moving. The surprise then increases when the ball enters the frame and starts accelerating.
We notice that when the ball is supposed to reappear, the model is not more surprised for the impossible than possible videos. This suggests that while the model has a long enough context to remember the object, looking almost 200 frames in the past is too difficult.

Nonetheless, while we see variations in surprise due to the seed in shared sections of the videos for impossible and possible ones, they are minor compared to the more general trends present throughout the videos. This suggests that models have some level of robustness to this natural noise.

\clearpage
\subsection{Generations from Cosmos}

\begin{figure}[tbp]
    \centering
    \includegraphics[width=1\linewidth]{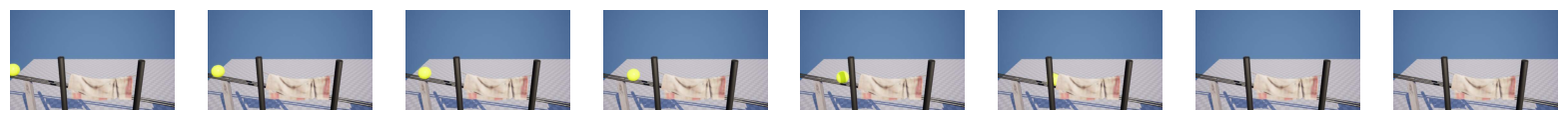}
    \includegraphics[width=1\linewidth]{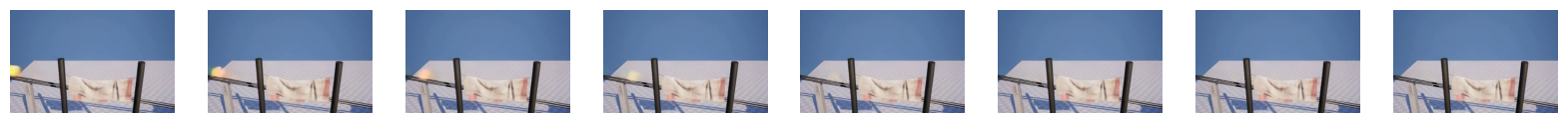}
    \includegraphics[width=1\linewidth]{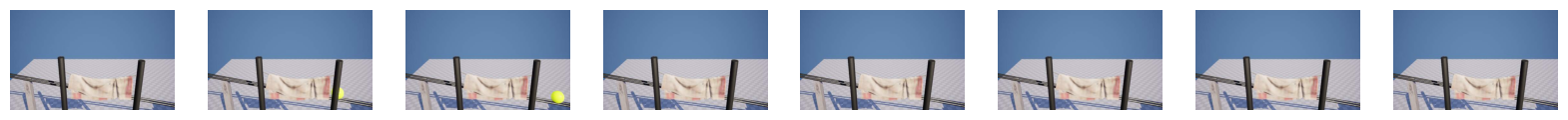}
    \includegraphics[width=1\linewidth]{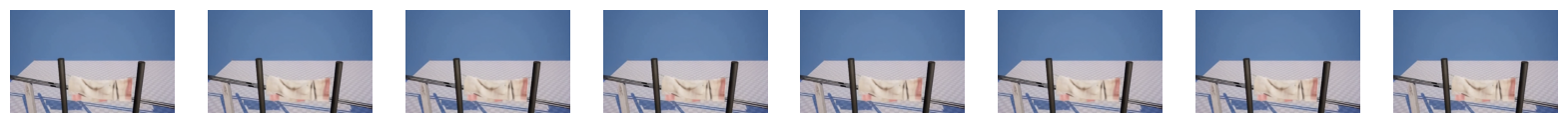}
    \caption{Top row: Context video given to Cosmos. 2nd row: Reconstruction of the encoded context. 3rd row: ground truth prediction. 4th row: Cosmos prediction. (2fps)}
    \label{fig:cosmos1}
\end{figure}

\begin{figure}[tbp]
    \centering
    \includegraphics[width=1\linewidth]{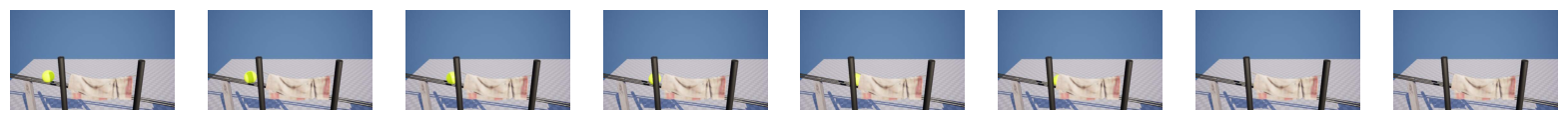}
    \includegraphics[width=1\linewidth]{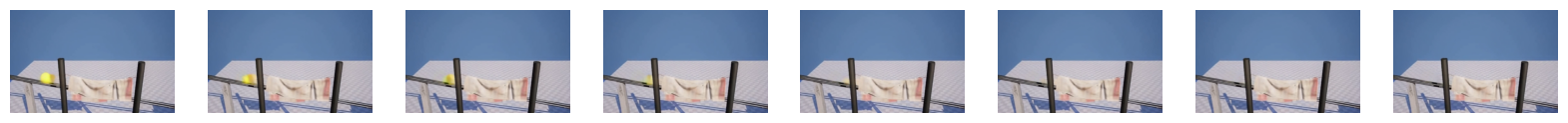}
    \includegraphics[width=1\linewidth]{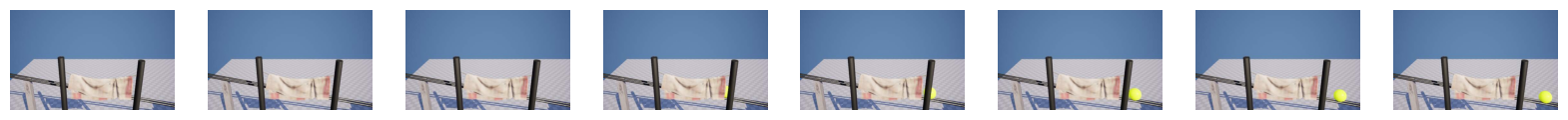}
    \includegraphics[width=1\linewidth]{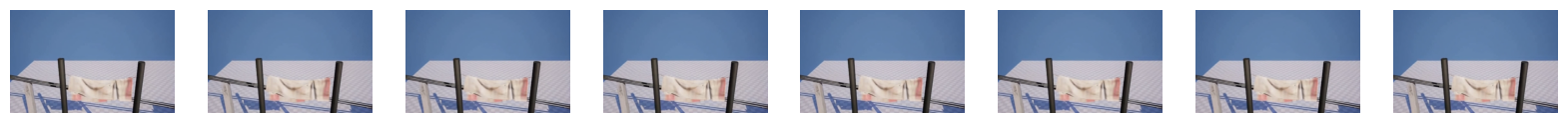}
    \caption{Top row: Context video given to Cosmos. 2nd row: Reconstruction of the encoded context. 3rd row: ground truth prediction. 4th row: Cosmos prediction. (6fps)}
    \label{fig:cosmos2}
\end{figure}

We perform a similar analysis with Cosmos, which has the advantage of allowing us to visualize the reconstructions of both the context, and predictions.

We visualize the predictions of Cosmos in figures~\ref{fig:cosmos1}~and~\ref{fig:cosmos2} on a video from the debug set. Looking at the context reconstructions in figure~\ref{fig:cosmos1}, we can see that the ball is not properly reconstructed after a few frames, even if it is still visible in the original context. We can hypothesize that the small size of the objects leads to it being forgotten by the video tokenizer, but this remains speculative. In figure~\ref{fig:cosmos2} this behavior is not observed as strongly, however we still see the object start to vanish as it approaches the occluder.

Looking at the prediction, we do not see the ball reappear in either setting. This corroborates the poor downstream performance we observe in the main results.
One thing to not is that in order to accommodate the context length of 9 that the model uses, we need to drastically reduce the framerate of the video. Combined with data that is out of distribution for the model this may lead to suboptimal performance. Nonetheless, even on this debug example we can see the model struggle.

\end{document}